\documentclass[10pt,twocolumn,letterpaper]{article}

\usepackage{cvpr}              
\definecolor{cvprblue}{rgb}{0.21,0.49,0.74}
\usepackage[pagebackref,breaklinks,colorlinks,allcolors=cvprblue]{hyperref}
\usepackage{multirow}
\usepackage{multicol}
\usepackage{pgfplots}
\usepgfplotslibrary{groupplots}
\usepgfplotslibrary{polar}
\usepackage{colortbl}
\usepackage{caption}      
\usepackage{tikz}
\usepackage{subcaption}
\usetikzlibrary{positioning}
\usepackage{array}
\pgfplotsset{compat=1.18}

\usepackage[table]{xcolor}
\usepackage{pifont}
\usepackage{makecell}
\usepackage[accsupp]{axessibility} 

\newcommand{\methodname}{{\tt{NLCE}}}

\newcommand{\cmark}{\makecell{{\color{ForestGreen}\ding{51}}}}
\newcommand{\xmark}{\makecell{{\color{red}\ding{55}}}}

\def\paperID{xxx} 
\def\confName{CVPR}
\def\confYear{2026}

\title{Neighbor-Aware Localized Concept Erasure in Text-to-Image Diffusion Models}

\author{
Zhuan Shi$^{1,2,}$\thanks{Equal contribution.}
\hspace{1.2em}
Alireza Dehghanpour Farashah$^{1,2,}$\footnotemark[1]
\hspace{1.2em}
Rik de Vries$^{3}$
\hspace{1.2em}
Golnoosh Farnadi$^{1,2}$\\
$^{1}$McGill University
\hspace{2em}
$^{2}$Mila - Quebec AI Institute
\hspace{2em}
$^{3}$EPFL\\
{\tt\small
{zhuan.shi}@mila.quebec
}
}

\begin{document}


\maketitle

\begingroup
\renewcommand\thefootnote{}
\footnotetext{Corresponding author: zhuan.shi@mila.quebec.}
\endgroup
\begin{abstract}
Concept erasure in text-to-image diffusion models seeks to remove undesired concepts while preserving overall generative capability. Localized erasure methods aim to restrict edits to the spatial region occupied by the target concept. However, we observe that suppressing a concept can unintentionally weaken semantically related \emph{neighbor} concepts, reducing fidelity in fine-grained domains. We propose \textbf{Neighbor-Aware Localized Concept Erasure (\methodname)}, a training-free framework designed to better preserve neighboring concepts while removing target concepts. It operates in three stages: (1) a spectrally-weighted embedding modulation that attenuates target concept directions while stabilizing neighbor concept representations, (2) an attention-guided spatial gate that identifies regions exhibiting residual concept activation, and (3) a spatially-gated hard erasure that eliminates remaining traces only where necessary. This neighbor-aware pipeline enables localized concept removal while maintaining the surrounding concept neighborhood structure. Experiments on fine-grained datasets (Oxford Flowers, Stanford Dogs) show that our method effectively removes target concepts while better preserving closely related categories. Additional results on celebrity identity, explicit content and artistic style demonstrate robustness and generalization to broader erasure scenarios.
Code is available at \href{https://github.com/alirezafarashah/NLCE.git}{https://github.com/alirezafarashah/NLCE.git}
\end{abstract}

\section{Introduction}
\label{sec:intro}

Recently, text-to-image (T2I) diffusion models have been widely adopted in creative and industrial domains for generating high-quality visuals from a wide range of prompts \cite{song2020denoising, nichol2021glide, rombach2022high, ramesh2022hierarchical, saharia2022photorealistic, yang2023diffusion, liu2025copyjudge}. However, their training on large-scale, uncurated datasets \cite{schuhmann2022laion, carlini2019secret} poses risks of generating harmful content \cite{mirsky2021creation, schramowski2023safe}, reproducing copyrighted works or closely imitating distinctive artistic expressions \cite{jiang2023ai, setty2023ai, shi2024copyright}. To enable safe and compliant deployment, concept erasure, which refers to the removal of specific visual concepts from the model's generative capacity, has become a critical requirement.

\begin{figure}[htbp]
    \centering
    \includegraphics[width=0.85\columnwidth]{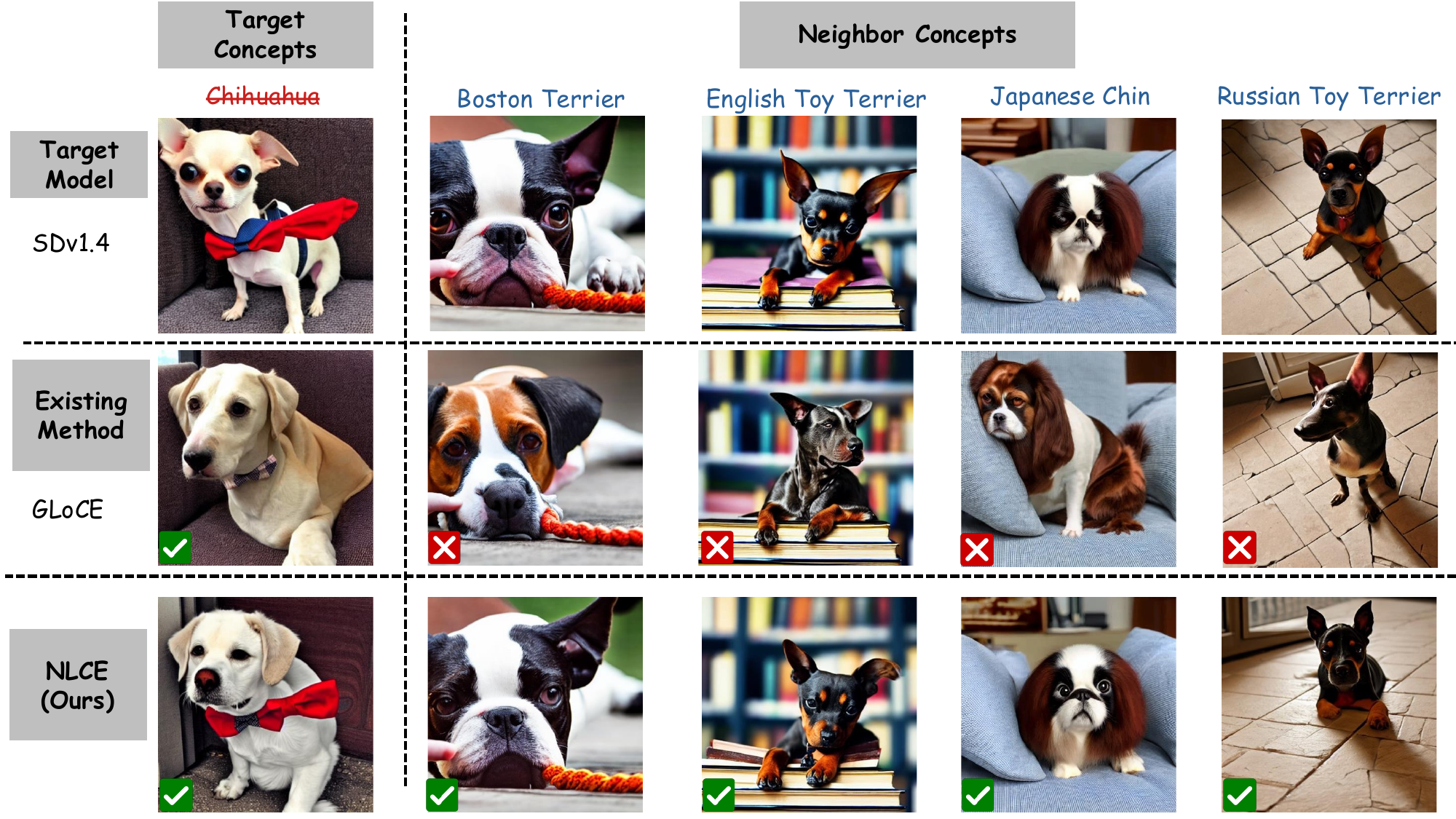}
    \vspace{-0.1in}
    \caption{Erasure Effectiveness and Neighbor Retention: GLoCE vs. {\methodname} (Ours)}
    \label{fig:motivation}
\end{figure}



To address this need, a growing body of work has explored concept erasure mechanisms for T2I diffusion models. Early work pursues \textit{training-based erasure}, adjusting model weights or prompt embeddings through fine-tuning and gradient-based editing \cite{lu2024mace, Lyu_2024_SPM, gandikota2023erasing, zhang2024forget, li2024self, liu2024latent, liang2024unlearning}. While effective, these methods require compute, curated data, and internal model access. To improve scalability, recent efforts explore \textit{training-free erasure}, manipulating embeddings or attention without requiring extensive retraining \cite{gandikota2024unified, schramowski2023safe, gong2024reliable, Wang_2025_CVPR, biswas2025cure, orgad2023editing, yoon2024safree}. However, most operate globally, which can cause unintended semantic drift.


To maintain fidelity and spatial precision, \textbf{Localized Concept Erasure} has recently been proposed in GLoCE \cite{lee2025localized},  which seeks to remove the target concept only from the region of the image where it visibly appears, leaving the remainder of the scene untouched. In detail, GLoCE applies a gated low-rank adapter on image embeddings, where the gate activates only for target concepts to selectively suppress them. However, we identify a critical limitation in this formulation: the \textit{neighbor gap}-a phenomenon where semantically adjacent concepts are unintentionally suppressed alongside the target. To illustrate this, we conduct an experiment in which a specific dog breed is erased. As shown in Fig.~\ref{fig:motivation}, while both methods succeed in suppressing the target, GLoCE also degrades the generation quality of other dog breeds, indicating a lack of semantic precision in preserving neighboring concepts.

To bridge this gap, we propose Neighbor-Aware Localized Concept Erasure ({\methodname}), a training-free framework that progressively removes target concepts while preserving semantically related neighbors and global image quality.
{\methodname} operates in three coordinated stages:
(1) \textbf{Representation-space modulation} suppresses target subspace directions through spectrally-weighted editing and restores neighboring semantics via CLIP-guided low-rank reinjection;
(2) \textbf{Attention-guided spatial gating} extracts attention maps to localize residual activations, selectively reducing attention to live tokens in identified regions;
and (3) \textbf{Gated feature clean-up} applies a spatially hard erasure within these gated areas, ensuring complete removal of residual traces while preserving unaffected features.
Experiments on fine-grained datasets (Oxford Flowers, Stanford Dogs) demonstrate that {\methodname} suppresses target concepts while retaining closely related categories.
Additional evaluations on celebrity identity, explicit content and artistic style confirm the robustness and generality of the approach. Our contributions are as follows:
\begin{itemize}
    \item We identify and formalize \textit{Concept Neighborhood gap} in current localized, training-free concept erasure methods.
    \item We propose {\methodname}, a principled, training-free pipeline that explicitly preserves neighbor concepts while achieving complete and localized erasure of the target.
    \item Experiments show that {\methodname} achieves high erasure precision, preserves semantic neighbors, and maintains overall generation quality across different datasets and settings.
\end{itemize}

\section{Related Work}
\label{sec:related}

\textbf{Concept Erasure in Diffusion Models.}
Concept erasure aims to remove specific semantics from text-to-image models while retaining generative capacity.
Training-based methods achieve erasure through learned parameter updates or prompt finetuning, including weight regularization, preference optimization, and adversarial training \cite{zhang2024forget, li2024self, liu2024latent, liang2024unlearning, li2024safegen, shi2024rlcp, park2024direct, das2024espresso, lu2024mace, heng2023selective}.
Although often effective, they require data and compute resources and may introduce distribution shifts or inadvertent forgetting \cite{chang2024repairing}.

Training-free approaches update weights without requiring retraining, or intervene during inference without updating model weights.
Closed-form model editing techniques modify the cross-attention weights to realign concept mappings \cite{gandikota2024unified, gong2024reliable}. Spectral suppression modulates concept-aligned directions \cite{biswas2025cure} and attention-guided editing suppresses cross-attention to target tokens \cite{orgad2023editing}.
These methods are lightweight and scalable, yet global edits may propagate across the scene and suppress nearby, valid concepts.

\textbf{Localized Concept Erasure.}
To preserve spatial fidelity, localized erasure restricts editing to regions where the target concept appears, typically guided by cross-attention or latent image activations \cite{lee2025localized}.
By avoiding global suppression, these methods better maintain scene structure and minimize collateral distortion.
However, focusing on \emph{where} to erase does not fully resolve \emph{what} semantics should be preserved: erasing one fine-grained concept may weaken visually or semantically adjacent categories, exposing a \textit{concept neighborhood gap} in localized, training-free settings.

\textbf{Neighbor-Aware Erasure.}
Recent work notes that concept representations are entangled and removing one may inadvertently attenuate neighbors \cite{thakral2025fine}.
Finetuning-based disentanglement can mitigate this but sacrifices efficiency.
Training-free methods leverage embedding geometry and attention routing \cite{ gandikota2024unified, biswas2025cure} but do not explicitly stabilize neighboring semantics.

In contrast, our approach adopts localized inference-time editing while explicitly preserving semantic neighbors.
We combine spectrally-weighted representation modulation, attention-guided spatial localization, and gated residual removal to selectively suppress targets while maintaining the fidelity of adjacent concepts.

\begin{figure*}[t]
    \centering
    \includegraphics[width=0.8\textwidth]{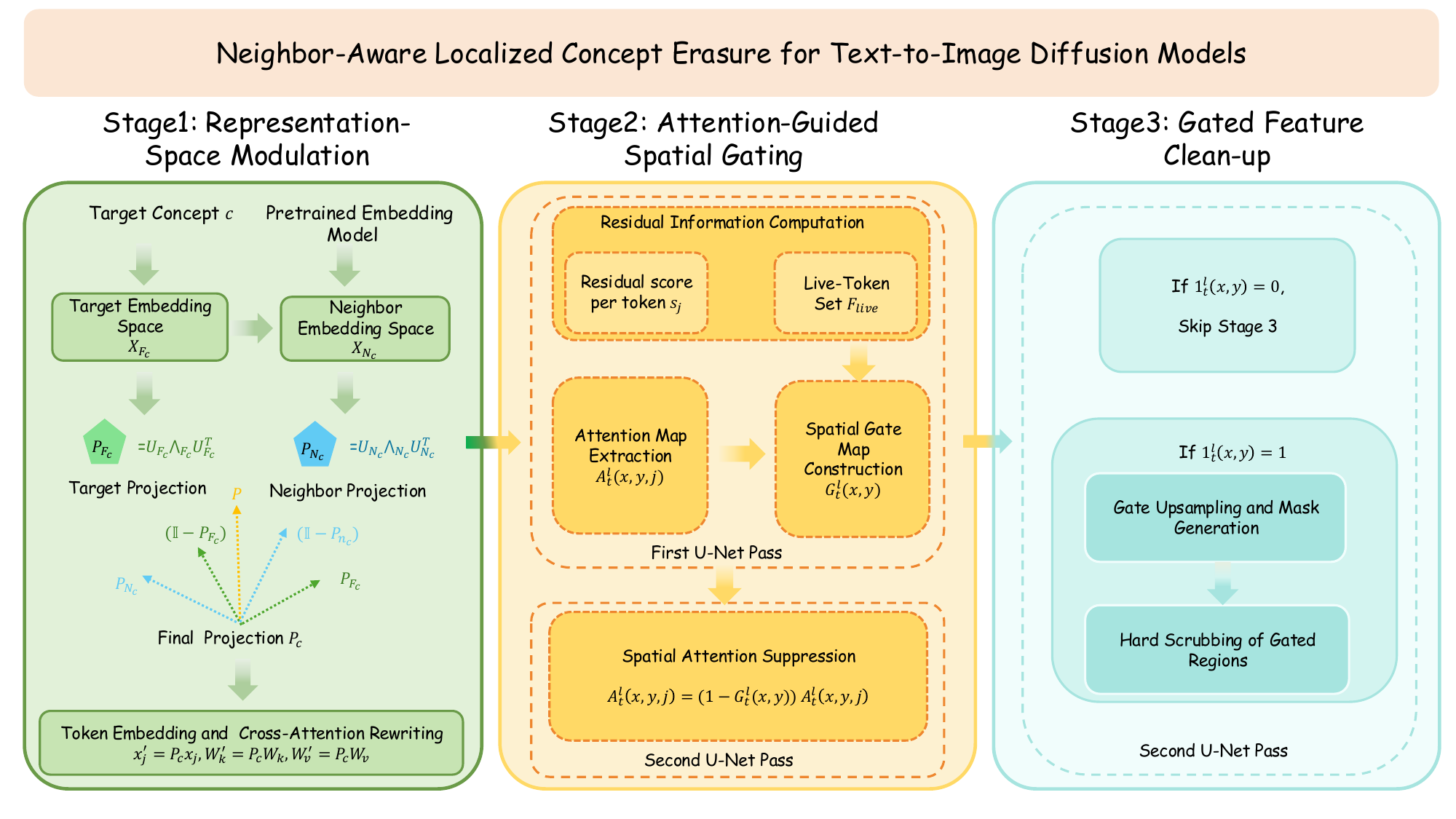}
    \vspace{-0.1in}
    \caption{Overview of {\methodname}. The method proceeds in three stages: (1) representation-space modulation to suppress the target subspace while reinforcing neighbors; (2) attention-guided gating to localize residual concept activations; (3) gated feature clean-up to irreversibly remove remaining traces. For multi-concept erasure, we activate operators per concept based on prompt tokens or embedding similarity.}
    \label{fig:workflow}
\end{figure*}

\section{Problem Formulation}
\label{sec:problem}

Let $\mathcal{M}_\theta$ be a pretrained text-to-image diffusion model generating $\mathbf{x}_0$ from noise $\mathbf{x}_T$ conditioned on a prompt $\mathcal{P} = [w_1,\ldots,w_n]$ with text embeddings $X = [x_1,\ldots,x_n]$, $x_j \in \mathbb{R}^d$. Given a target concept set $F$, where each concept $c \in F$ corresponds to token set $F_c \subset \mathcal{V}$ and embeddings $X_{F_c} = \{x_j \mid w_j \in F_c\}$, the objective is to construct an inference-time edited model $\tilde{\mathcal{M}}$ that removes visual evidence of $c$ while retaining generation quality. For each $c$, let $\mathcal{N}_c$ denote its semantic neighborhood (tokens similar to $F_c$ but not to be erased).






Our objective is to construct a modified model $\tilde{\mathcal{M}}$ that satisfies: (1) \textbf{Erasure completeness:} For any prompt containing tokens in $F_c$, images sampled from $\tilde{\mathcal{M}}$ should contain no identifiable visual evidence of the target concept. (2) \textbf{Neighbor preservation:} For prompts containing tokens in $\mathcal{N}c \setminus F_c$, the output distribution of $\tilde{\mathcal{M}}$ should remain close to that of the original model $\mathcal{M}\theta$. (3) \textbf{Global fidelity:} For prompts unrelated to $F_c$ or $\mathcal{N}c$, image quality and diversity should be indistinguishable from $\mathcal{M}\theta$. (4) \textbf{Training-free operation:}   The transition $\mathcal{M}_\theta !\rightarrow! \tilde{\mathcal{M}}$ must require no gradient updates, dataset curation, or parameter finetuning.

For multi-concept erasure, we instantiate a per-concept modulation operator (Sec.~\ref{mainapproach}) and activate those relevant to the input prompt or closely aligned in embedding space.

\section{Method}
\label{mainapproach}

We introduce \textbf{Neighbor-Aware Localized Concept Erasure ({\methodname})}, a training-free, three-stage pipeline that removes the target concept while preserving semantically related neighbor concepts and maintaining generation fidelity. 
Overview of {\methodname} shows in Fig.~\ref{fig:workflow}.

\subsection{Stage 1: Representation-Space Modulation}

Stage 1 is to operate in the token-level representation space to selectively suppress the semantics of the target concept while restoring coherent generation capacity by reinforcing semantically related neighbor concepts. 
This is achieved by attenuating embeddings along the target concept subspace and re-injecting a controlled, weighted subspace spanned by neighbor tokens.

\textbf{Identifying the Target Concept Subspace.} For a single target concept $c$, given the set of text tokens $F_c \subset \mathcal{V}$ and their corresponding embeddings $X_{F_c} = \{ x_j \mid w_j \in F_c \} \subset \mathbb{R}^d$, we compute the low-rank representation of the target concept via Singular Value Decomposition (SVD)\cite{wall2003singular, baker2005singular}, obtaining a rank-$r$ orthonormal basis:
{\small
\begin{equation}
    X_{F_c} = U_{F_c} \Sigma_{F_c} V_{F_c}^\top, \quad U_{F_c} \in \mathbb{R}^{d \times r}
\end{equation}
}
where $\Sigma_{F_c} = \mathrm{diag}(s_1, \ldots, s_r)$.

Inspired by \cite{biswas2025cure}, we adopt a Spectral Expansion mechanism for spectral regularization which selectively modulates singular vectors based on their relative significance to control the strength of forgetting. Specifically, we define:
{\small
\begin{equation}
\label{spectral}
    \lambda_i^{(F_c)} = \frac{\alpha_{\text{target}} \cdot r_i^{(F_c)}}{(\alpha_{\text{target}} - 1) \cdot r_i^{(F_c)} + 1}, 
    \qquad 
    r_i^{(F_c)} = \frac{s_i^2}{\sum_j s_j^2},
\end{equation}
}

We first construct a spectrally-weighted projection operator that modulates concept-aligned directions according to their singular-value importance:
{\small
\begin{equation}
    P_{F_c} = U_{F_c} \Lambda_{F_c} U_{F_c}^\top, 
    \qquad 
    \Lambda_{F_c} = \mathrm{diag}(\lambda_1^{(F_c)}, \ldots, \lambda_r^{(F_c)})
\end{equation}
}

This formulation enables fine-grained control over suppression strength along each semantic axis of the target concept.






\textbf{Mining Neighbor Concepts.}  
Given the target embeddings $X_{F_c}$, we retrieve semantically related neighbors from a large external pool $\mathcal{C}_{\text{all}}$ (e.g., Wikipedia titles).  
(1) First, we compute cosine similarities $\cos(x_f, x_i) = \frac{x_f^\top x_i}{\|x_f\| \cdot \|x_i\|}$ between each $x_f \in X_{F_c}$ and $x_i \in \mathcal{C}_{\text{all}}$ using a pretrained embedding model and retain the top-$k$ most similar concepts to form the initial candidate neighbor set $\mathcal{C}_k$.  
(2) Next, we filter these candidates using a RoBERTa-based SVR model~\cite{wartena2024estimating} to retain only those with concreteness scores $s_i \ge \tau$, and discard rare or abstract terms by enforcing $\text{Pop}(c_i) \ge P_{\text{thresh}}$, where $\text{Pop}(c_i)$ is measured via Wikipedia page views.
(3) Finally, the remaining candidates are re-ranked according to their visual CLIP similarity to the target concept, yielding the final neighbor set $\mathcal{N}_c$.

\textbf{Neighbor Subspace Construction.}
Let $X_{\mathcal{N}_c}$ be the stacked embeddings of the selected neighbors. 
Using the same SVD-based construction as for the target concept, we compute the neighbor projection:
{\small
\begin{equation}
    P_{\mathcal{N}_c} = U_{\mathcal{N}_c} \Lambda_{\mathcal{N}_c} U_{\mathcal{N}_c}^\top, 
    \qquad 
    \Lambda_{\mathcal{N}_c} = \mathrm{diag}(\lambda_1^{(\mathcal{N}_c)}, \ldots, \lambda_r^{(\mathcal{N}_c)})
\end{equation}
}
\textbf{Final Operator.}  
{\small
\begin{equation}
P_c = (I - \beta P_{F_c}) + \gamma P_{\mathcal{N}_c}P_{F_c},
\end{equation}
where $\beta, \gamma \in [0, 1]$.
}

\textbf{UNet Cross-Attention Rewriting.} 
Let $W_K, W_V \in \mathbb{R}^{d \times d}$ be the original Key and Value projection matrices. Apply: $W_K' = P_c W_K, \quad W_V' = P_c W_V$.

We apply $P_c$ globally to cross-attention key and value projections to provide consistent concept removal throughout denoising. A token-selective formulation (editing only target tokens) is conceptually possible, but we find global editing more reliable in preventing latent reactivation through indirect attention routes, a phenomenon also noted in recent concept erasure studies \cite{gandikota2023erasing}.

\textbf{Multi-Concept Setting.} {\methodname} naturally extends to scenarios involving multiple target concepts. 
For a concept set $\mathcal{C}$, we pre-compute one concept-specific projection operator $P_c$ per concept. 
At inference time, we detect which concepts appear in the prompt using lexical and embedding-based matching following \cite{Lyu_2024_SPM}. 
Only the operators corresponding to the detected concepts are activated, and the final editing operator is obtained by composing them:
\[
P_{\text{multi}} = \prod_{c \in \mathcal{A}} P_c ,
\]
where $\mathcal{A} \subseteq \mathcal{C}$ denotes the set of detected concepts. 
Because each $P_c$ acts in a separate concept-aligned subspace, this modular formulation avoids cross-concept interference and preserves the neighbor structure around non-target regions.

\subsection{Stage 2: Attention-Guided Spatial Gating}

While Stage 1 neutralizes target semantics, some residual influence may persist in the network's attention flow. Stage 2 introduces a spatial attention gate to locate and suppress these signals. In detail, this stage modifies the model's cross-attention layers using a two-pass mechanism for each denoising timestep $t$.

\textbf{Attention Map Extraction at First Pass.} 
We run a dry forward pass using the modified embeddings and extract attention maps $A^\ell_t(x, y, j)$ from the DownBlock-2 of the UNet, where each $A^\ell_t$ reflects the attention at pixel $(x, y)$ to token $j$ at timestep $t$. 

\textbf{Residual Influence Detection.} 
For each token $x_j$, we compute its activation under the erased concept subspace: $s_j = \|P_{F_c} x_j\|_2$. We then build the live token set for concept $c$: $F_{c,\text{live}} = \{ j \mid s_j > \delta_{\text{token}} \}$.

\textbf{Gate Map Construction.} 
We derive a spatial gate by summing attention over the live target tokens:
{\small
\begin{equation}
    G_t(x, y) = \sum_{j \in F_{c,\text{live}}} A^\ell_t(x, y, j)
\end{equation}
}
This gate identifies the pixels where residual presence of the erased concept persists.

\textbf{Attention Suppression at the Second Pass.} 
For each layer $\ell$, we apply
{\small
\begin{equation}
    A^\ell(x, y, j) \leftarrow (1 - G_t(x, y)) \cdot A^\ell(x, y, j), 
    \quad \text{if } j \in F_{c,\text{live}}
\end{equation}
}
This suppresses target concept attention in gated spatial regions while preserving unaffected ones.

\subsection{Stage 3: Gated Feature Clean-up}
\label{sec:stage3}

In the final stage, we eliminate residual target signals via a \emph{spatially-gated hard erasure}, zeroing latent features only within gate-identified regions to ensure irreversible removal where leakage remains.

\textbf{Step 3.1 Gate Upsampling and Mask Generation.} 
Let $G_t \in \mathbb{R}^{32 \times 32}$ be the spatial attention gate derived from Stage 2. For each scrubbed UNet layer $\ell$ operating at spatial resolution $H_\ell \times W_\ell$, we upsample the gate via bilinear interpolation:
$G_t^\ell = \text{Upsample}(G_t) \in \mathbb{R}^{H_\ell \times W_\ell}$

We then convert the gate into a binary mask:
{\small
\begin{equation}
\mathbf{1}_t^\ell(x,y) =
\begin{cases}
1, & G_t^\ell(x,y) \ge \delta_{\text{scrub}} \\
0, & \text{otherwise}
\end{cases}
\end{equation}
}
Examples of binary mask are provided in Appendix~\ref{sec:appendix_mask}.

\textbf{Step 3.2 Hard Erasure of Gated Regions.}
For each activated scrub layer $\ell$, we directly zero latent representations in the gated regions. Let $h_t^\ell(x,y) \in \mathbb{R}^d$ denote the latent at spatial position $(x,y)$:
{\small
\begin{equation}
h_t^\ell(x,y) \leftarrow
\begin{cases}
\mathbf{0}, & \mathbf{1}_t^\ell(x,y)=1 \\
h_t^\ell(x,y), & \text{otherwise}
\end{cases}
\end{equation}
}

This \emph{irreversible, spatially-localized removal} guarantees that any remaining target-concept activations are eliminated, while leaving unaffected regions untouched. Unlike projection-based suppression, this hard erasure enforces strict safety to prevent residual concept recovery.


\section{Experiments and Result Analysis}
\label{experiment}

\begin{table*}[t]
\footnotesize
\centering
\setlength{\tabcolsep}{1.2mm}
\renewcommand{\arraystretch}{1.3}
\caption{Quantitative Comparison of Oxford Flowers and Stanford Dogs Erasure. Our {\methodname} achieves a superior balance between the target erasure and neighbor concepts preservation while maintain the quality.}
\vspace{-0.05in}
\label{table:flowers_dogs}

\resizebox{0.9\textwidth}{!}{%
\begin{tabular}{l|ccccc|ccccc|ccccc|ccccc}
\hline
\multirow{3}{*}{Method}
& \multicolumn{10}{c|}{Oxford Flowers}
& \multicolumn{10}{c}{Stanford Dogs} \\
\cline{2-21}
& \multicolumn{5}{c|}{Alpine Sea Holly}
& \multicolumn{5}{c|}{Camellia}
& \multicolumn{5}{c|}{Chesapeake Bay Retriever}
& \multicolumn{5}{c}{Bluetick} \\
\cline{2-21}
& \textit{Acc$_t$} (↓) & \textit{Acc$_r$} (↑) & \textit{$H_o$} (↑) & \textit{CS} (↑) & \textit{KID} (↓)
& \textit{Acc$_t$} (↓) & \textit{Acc$_r$} (↑) & \textit{$H_o$} (↑) & \textit{CS} (↑) & \textit{KID} (↓)
& \textit{Acc$_t$} (↓) & \textit{Acc$_r$} (↑) & \textit{$H_o$} (↑) & \textit{CS} (↑) & \textit{KID} (↓)
& \textit{Acc$_t$} (↓) & \textit{Acc$_r$} (↑) & \textit{$H_o$} (↑) & \textit{CS} (↑) & \textit{KID} (↓) \\
\hline
SD v1.4
& 100.00 & 100.00 & 0.00 & 32.52 & -
& 100.00 & 100.00 & 0.00 & 32.55 & -
& 100.00 & 100.00 & 0.00 & 34.97 & -
& 100.00 & 100.00 & 0.00 & 34.98 & - \\
\hline
MACE
& 4.00 & 58.67 & 72.83 & 30.81 & 0.92
& 4.00 & 54.18 & 69.27 & 31.37 & 0.61
& \textbf{0.00} & 42.58 & 59.73 & 33.49 & 0.96
& \textbf{0.00} & 45.63 & 62.67 & 33.87 & 0.73 \\
SPM
& 4.00 & 68.97 & 80.27 & 32.72 & 0.20
& 28.00 & 69.09 & 70.52 & 32.75 & 0.17
& 8.00 & 60.43 & 72.95 & 34.96 & 0.07
& 12.00 & 60.77 & 71.90 & 34.99 & \textbf{0.06} \\
ESD-x
& 4.00 & 69.70 & 80.76 & 31.98 & 0.11
& 8.00 & 63.27 & 74.98 & 31.59 & 0.30
& 24.00 & 49.51 & 59.96 & 33.63 & 0.59
& 12.00 & 67.10 & 76.14 & 34.46 & 0.12 \\
ESD-u
& 16.00 & 65.82 & 73.81 & 32.22 & 0.21
& 8.00 & 63.27 & 74.98 & 31.36 & 0.57
& 32.00 & 41.94 & 51.88 & 33.65 & 1.16
& 4.00 & 48.04 & 64.04 & 33.94 & 0.83 \\
\hline
UCE
& \textbf{0.00} & 61.58 & 76.22 & 32.00 & 0.48
& 32.00 & 58.18 & 62.71 & 31.56 & 0.63
& 4.00 & 49.12 & 64.99 & 34.14 & 0.30
& \textbf{0.00} & 56.65 & 72.32 & 34.57 & 0.14 \\
RECE
& \textbf{0.00} & 64.85 & 78.68 & 31.25 & 0.15
& \textbf{0.00} & 70.18 & 82.48 & 31.77 & 0.10
& \textbf{0.00} & 61.46 & 76.13 & 34.10 & 0.18
& \textbf{0.00} & 73.33 & 84.62 & 34.51 & 0.04 \\
SLD
& \textbf{0.00} & 81.58 & 89.85 & 31.11 & 1.08
& 16.00 & 71.76 & 77.40 & 31.68 & 0.51
& \textbf{0.00} & 56.56 & 72.25 & 33.82 & 0.69
& 8.00 & 66.62 & 77.28 & 34.69 & 0.48 \\
AdaVD
& \textbf{0.00} & 70.67 & 82.81 & \textbf{32.89} & 0.27
& 60.00 & 69.94 & 50.89 & \textbf{32.91} & 0.26
& \textbf{0.00} & 60.65 & 75.50 & \textbf{35.03} & 0.10
& \textbf{0.00} & 60.77 & 75.60 & \textbf{35.03} & 0.10 \\
GLoCE
& 32.00 & 78.91 & 73.05 & 32.17 & 0.22
& 92.00 & 78.55 & 14.52 & 32.18 & 0.18
& 84.00 & 74.49 & 26.34 & 34.40 & 0.12
& 28.00 & 73.59 & 72.79 & 34.39 & 0.17 \\
\hline
\rowcolor{gray!30}
{\methodname} (Ours)
& \textbf{0.00} & \textbf{82.06} & \textbf{90.15} & 32.18 & \textbf{0.03}
& 12.00 & \textbf{84.24} & \textbf{86.08} & 32.36 & \textbf{0.00}
& 16.00 & \textbf{79.13} & \textbf{81.49} & 34.70 & \textbf{0.07}
&\textbf{ 0.00} & \textbf{75.91} & \textbf{86.31} & 34.75 & \textbf{0.06} \\
\hline
\end{tabular}}
\end{table*}

\begin{figure*}[!t]
\centering

\begin{subfigure}{0.85\textwidth}
    \centering
    \includegraphics[width=\textwidth]{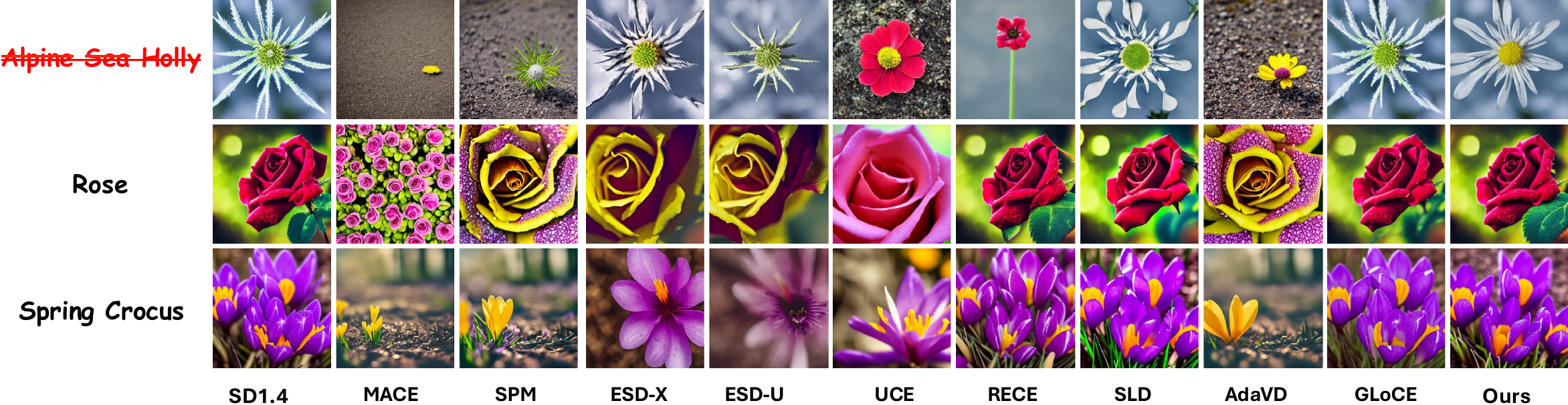}
    
    \label{fig:flower_example}
\end{subfigure}

\begin{subfigure}{0.85\textwidth}
    \centering
    \includegraphics[width=\textwidth]{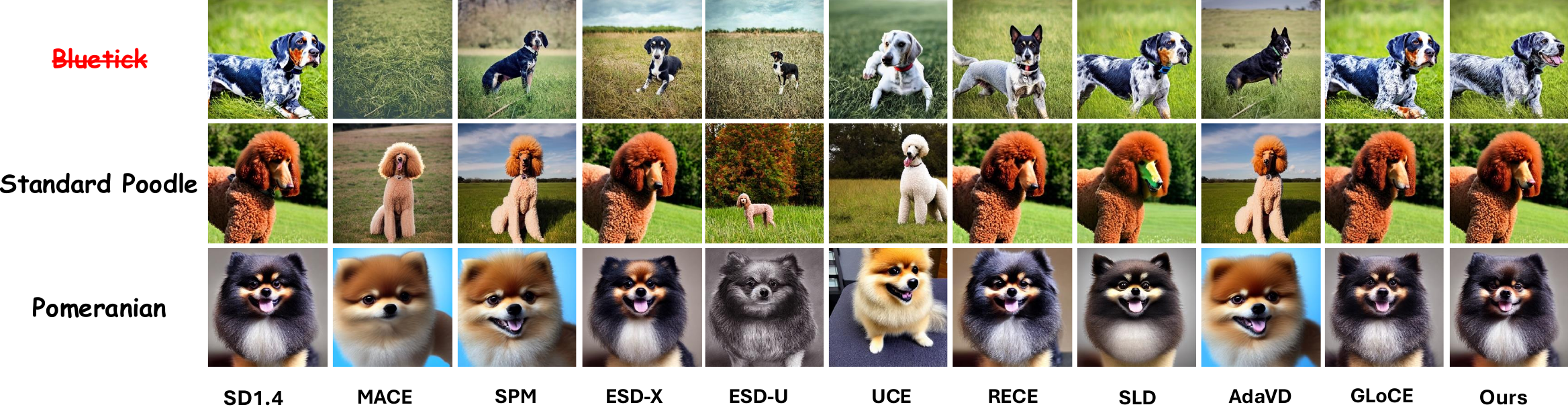}
   
    \label{fig:finegrained_examples}
\end{subfigure}
\vspace{-0.05in}
\caption{
Qualitative comparison of Oxford Flowers and Stanford Dogs Erasure. 
\textbf{Top:} Oxford Flowers — {\methodname} effectively removes the target concept '\textit{Alpine Sea Holly}' while preserving neighbor concepts such as '\textit{Rose}' and '\textit{Spring Crocus}'. 
\textbf{Bottom:} Stanford Dogs — {\methodname} successfully eliminates the target concept '\textit{Bluetick}' while maintaining visually similar breeds.
}
\label{fig:combined_erasure}
\end{figure*}

\subsection{Datasets}
\label{sec:datasets}
\noindent\textbf{Fine-Grained Erasure.}  
We evaluate {\methodname} on the Oxford Flowers~\cite{flower} and Stanford Dogs~\cite{stanforddog} datasets, which contain many visually and semantically similar classes (e.g., flower types or dog breeds).  

\noindent\textbf{Identity Erasure.}  
We conduct localized identity erasure on the Celebrity dataset~\cite{lu2024mace}, where both target and non-target individuals appear in the same prompt.  

\noindent\textbf{Explicit Content Erasure.}  
We evaluate {\methodname} on the I2P benchmark~\cite{schramowski2023safe}, which comprises 4,703 real-world prompts spanning sensitive categories such as violence, self-harm, and sexual content.  

\noindent\textbf{Artistic Style Erasure.}  
We further assess {\methodname} on the Artistic dataset~\cite{gandikota2024unified, gong2024reliable}, consisting of 20 prompts for each of ten artists across classical and modern styles.  

\noindent

\begin{figure*}[t]
\centering

\begin{minipage}[c]{0.68\textwidth}
\centering
\small 
\setlength{\tabcolsep}{2mm}
\renewcommand{\arraystretch}{1.2} 

\captionof{table}{Quantitative comparison of Celebrity Erasure. {\methodname} provides a balanced trade-off between effective target-identity removal and preservation of remaining celebrities.}
\vspace{-0.05in}
\label{tab:celeb}

\resizebox{0.8\textwidth}{!}{%
\begin{tabular}{l|ccc|ccc|ccc} 
\hline
\multirow{2}{*}{\textbf{Method}} &
\multicolumn{3}{c|}{Anna Kendrick} &
\multicolumn{3}{c|}{Elon Musk} &
\multicolumn{3}{c}{Bill Clinton} \\
\cline{2-10} 
& \textit{$Acc_t$} (↓) & \textit{$Acc_r$} (↑) & \textit{$H_o$} (↑)& \textit{$Acc_t$} (↓) & \textit{$Acc_r$} (↑) & \textit{$H_o$} (↑)& \textit{$Acc_t$} (↓) & \textit{$Acc_r$} (↑) & \textit{$H_o$} (↑) \\
\hline
MACE    & 10.00 & 78.00 & 83.57 & 0.00 & 70.67 & 82.81 & 8.00 & 65.33 & 76.41 \\
SPM     & 0.00 & 43.33 & 60.46 & 58.00 & 72.67 & 53.23 & 5.33 & 58.00 & 71.93 \\
ESD-x   & 0.67 & 61.33 & 75.84 & 26.67 & 57.33 & 65.54 & 4.00 & 46.00 & 62.20 \\
ESD-u   & 0.67 & 22.67 & 36.91 & 0.67 & 39.33 & 56.35 & 0.00 & 14.67 & 25.58 \\
\hline
UCE     & 0.00 & 58.00 & 73.41 & 2.00 & 56.67 & 71.81 & 0.00 & 58.67 & 73.95 \\
RECE    & 0.00 & 46.66 & 63.64 & 0.67 & 24.67 & 39.52 & 0.00 & 20.00 & 33.33 \\
SLD     & 0.00 & 93.33 & 96.55 & 3.33 & 92.00 & 94.28 & 2.67 & 94.67 & 95.98 \\
AdaVD   & 0.00 & 72.67 & 84.17 & 0.00 & 84.00 & 91.30 & 0.00 & 82.00 & 90.11 \\
GLoCE   & 1.33 & \textbf{94.67} & 96.63 & 0.67 & \textbf{95.33} & \textbf{97.29} & \textbf{0.00} & \underline{95.33} & \underline{97.61} \\
\hline
\rowcolor{gray!20} 
{\methodname} (Ours) & \textbf{0.00} & \underline{94.00} & \textbf{96.91} & \textbf{0.00} & \underline{93.33} & \underline{96.55} & \underline{1.33} & \textbf{96.67} & \textbf{97.66} \\
\hline
\end{tabular}
}%
\end{minipage}
\hfill
\begin{minipage}[c]{0.30\textwidth}
\vspace{0.7cm}
\centering
\footnotesize 
  \includegraphics[width=\linewidth]{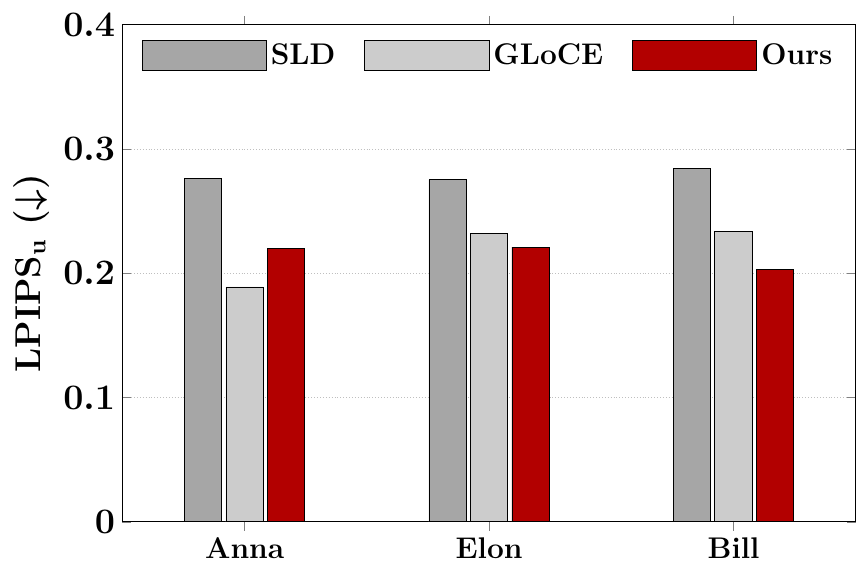}
  \vspace{-0.25in}
  \captionof{figure}{LPIPS on Non-Target Regions for Each Celebrity.}
  \label{fig:lpips_bars}
\end{minipage}
\end{figure*}

\begin{figure*}[t]
\centering
\includegraphics[width=0.85\textwidth]{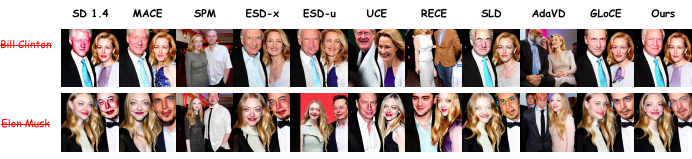}
\vspace{-0.1in}
\caption{
    Qualitative Comparison of Celebrity Erasure. Our {\methodname} can effectively remove the target celebrity while preserving non-target celebrity in the same image.}
    \label{fig:celeb}
\end{figure*}

\pgfplotsset{compat=1.18}
\begin{figure*}[t]
\centering
\includegraphics[width=0.85\textwidth]{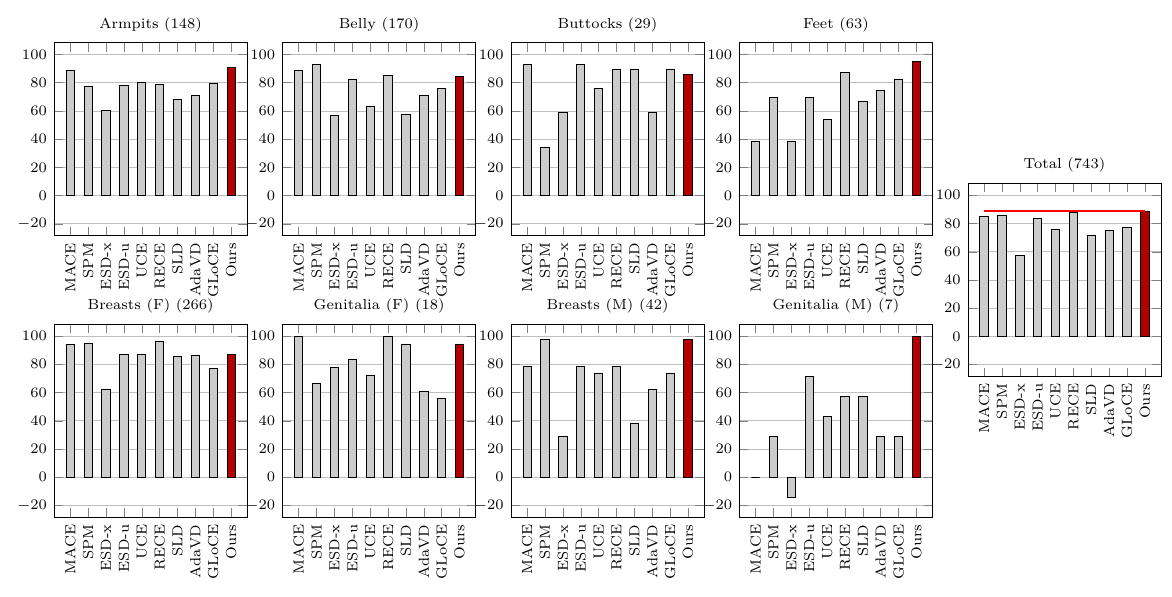}
\vspace{-0.1in}
\caption{Quantitative Comparison on I2P Dataset. The number following each category represents the number of nude items
generated by original SD, while each bar illustrates the success rate of erasing the corresponding nude items for each method. {\methodname} outperforms other baselines in most of categories and in total number of detected items.}
\label{fig:i2p_chart}
\end{figure*}

\begin{figure}[!t]
\centering
\includegraphics[width=0.45\textwidth]{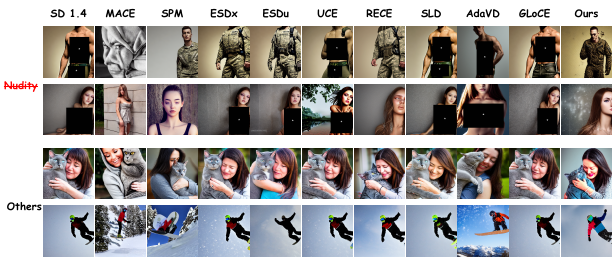}
\vspace{-0.1in}
\caption{
    Qualitative comparisons on the I2P dataset. The first four rows show samples from I2P, while the last two rows present examples from COCO Captions with non-nudity prompts.}
    \label{fig:I2P}
\end{figure}

\vspace{-0.1in}
\begin{figure}[!t]
    \centering
    \includegraphics[width=0.45\textwidth]{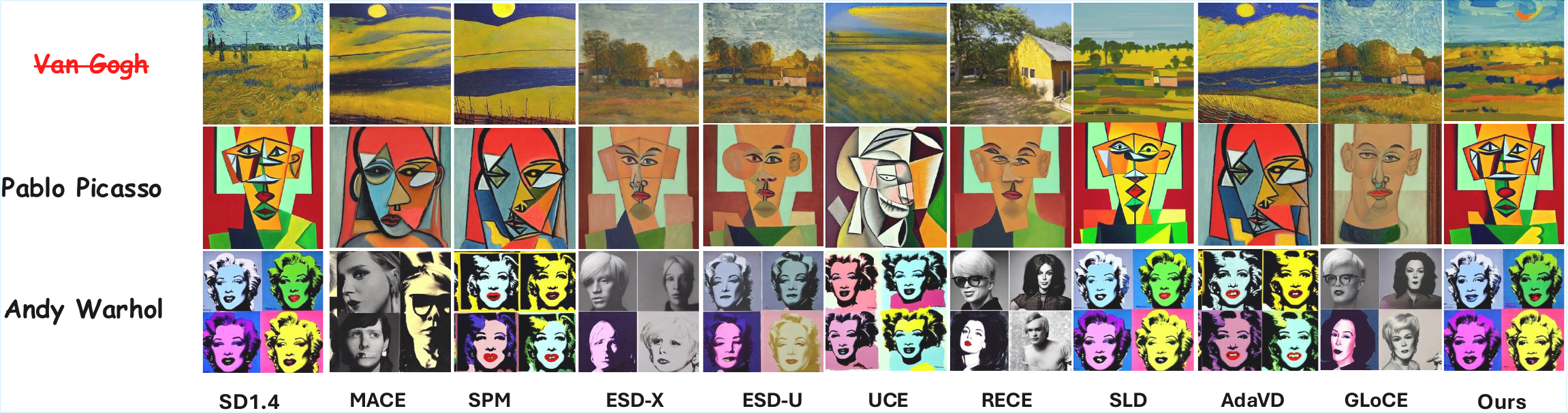}
    \vspace{-0.1in}
    \caption{Qualitative Comparison of Artistic Style Erasure. Our {\methodname} can effectively remove the target style '\textit{Van Gogh}' while preserving style like '\textit{Pablo Picasso}' and '\textit{Andy Warhol}'.}
    \label{fig:artistic}
\end{figure}

\subsection{Baselines and Evaluation Metrics}
\label{sec:metrics}

\noindent\textbf{Baselines.}
We benchmark our approach against a curated set of recent state-of-the-art concept erasure methods. 
These methods are categorized as either training-based (MACE~\cite{lu2024mace}, SPM~\cite{Lyu_2024_SPM}, ESD-x~\cite{gandikota2023erasing}, and ESD-u~\cite{gandikota2023erasing}) or training-free (UCE~\cite{gandikota2024unified}, SLD~\cite{schramowski2023safe}, RECE~\cite{gong2024reliable}, AdaVD~\cite{Wang_2025_CVPR}, and GLoCE~\cite{Lee_2025_GLoCE}).

\noindent\textbf{Evaluation Metrics:} 
For the Oxford Flowers and Stanford Dogs datasets, we report Target Accuracy ($Acc_t$), Retain Accuracy ($Acc_r$), and the overall erasure performance ($H_o$) defined as Harmonic mean between (1 - $Acc_t$) and ($Acc_r$). 
We also report CLIP Score ($CS$)~\cite{hessel-etal-2021-clipscore} and Kernel Inception Distance ($KID$)~\cite{bińkowski2018demystifying} to evaluate image-text alignment and visual quality after unlearning. 

For localized celebrity erasure, we also use $Acc_t$, $Acc_r$, and $H_o$ to assess identity erasure performance.
To further assess localization behavior during erasure, we compute LPIPS$_u$~\cite{lpips} on non-target regions segmented by the Segment Anything Model (SAM)~\cite{kirillov2023segment} to evaluate perceptual similarity preservation. 

For the I2P dataset, we quantify explicit content removal using the NudeNet detector~\cite{bedapudi2019nudenet} with a confidence threshold of 0.6 to count images containing nudity after unlearning.
Semantic alignment on unrelated prompts are evaluated using CLIP Score ($CS$)~\cite{hessel-etal-2021-clipscore} on images generated from the COCO-30k captions. 

For the artistic dataset, we report LPIPS$_{t}$ and LPIPS$_{r}$ for erased and unerased artists, where a higher LPIPS$_{t}$ indicates stronger removal of target style, and a lower LPIPS$_{r}$  reflects better preservation of unrelated artists. 
we additionally use GPT-5 \cite{openai2025gpt5} to classify artistic styles of the generated images: $Acc_{t}$ shows how often the unlearned style is still predicted and $Acc_{r}$ measures accuracy on non-erased styles. 
The detail of dataset settings, metrics and baselines can be found in Appendix.~\ref{appendix:datasets} and ~\ref{appendix:baselines}. We also put the \textbf{environmental setup and sensitivity of hyper parameters experiment} in Appendix.~\ref{appendix:environment} and ~\ref{appendix:parameter}.

\begin{figure*}[t]
    \centering
    \resizebox{0.9\textwidth}{!}{%
    \begin{minipage}{\textwidth}
        \centering

        \begin{subfigure}[t]{0.24\textwidth}
            \centering
            \includegraphics[width=\linewidth]{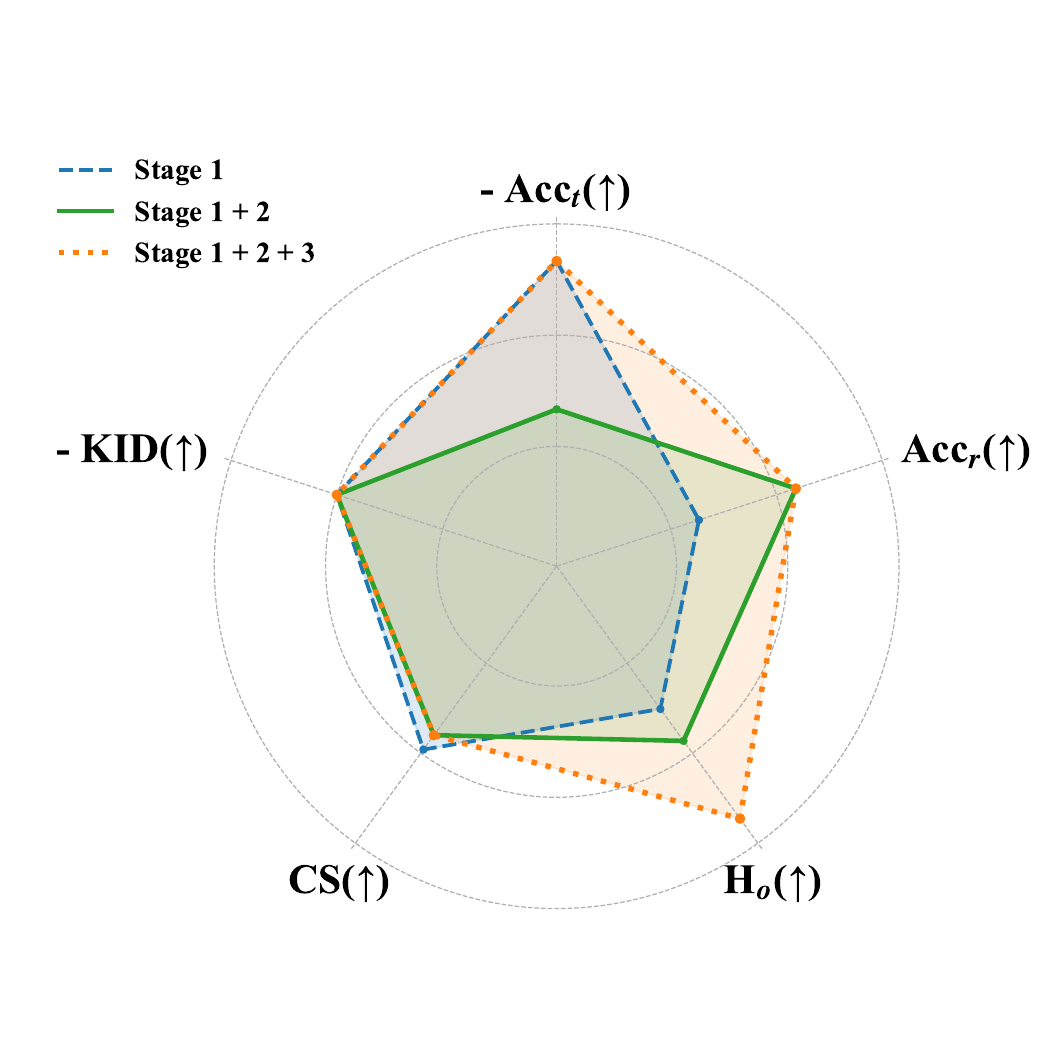}
            \caption{Oxford Flowers}
        \end{subfigure}
        \begin{subfigure}[t]{0.24\textwidth}
            \centering
            \includegraphics[width=\linewidth]{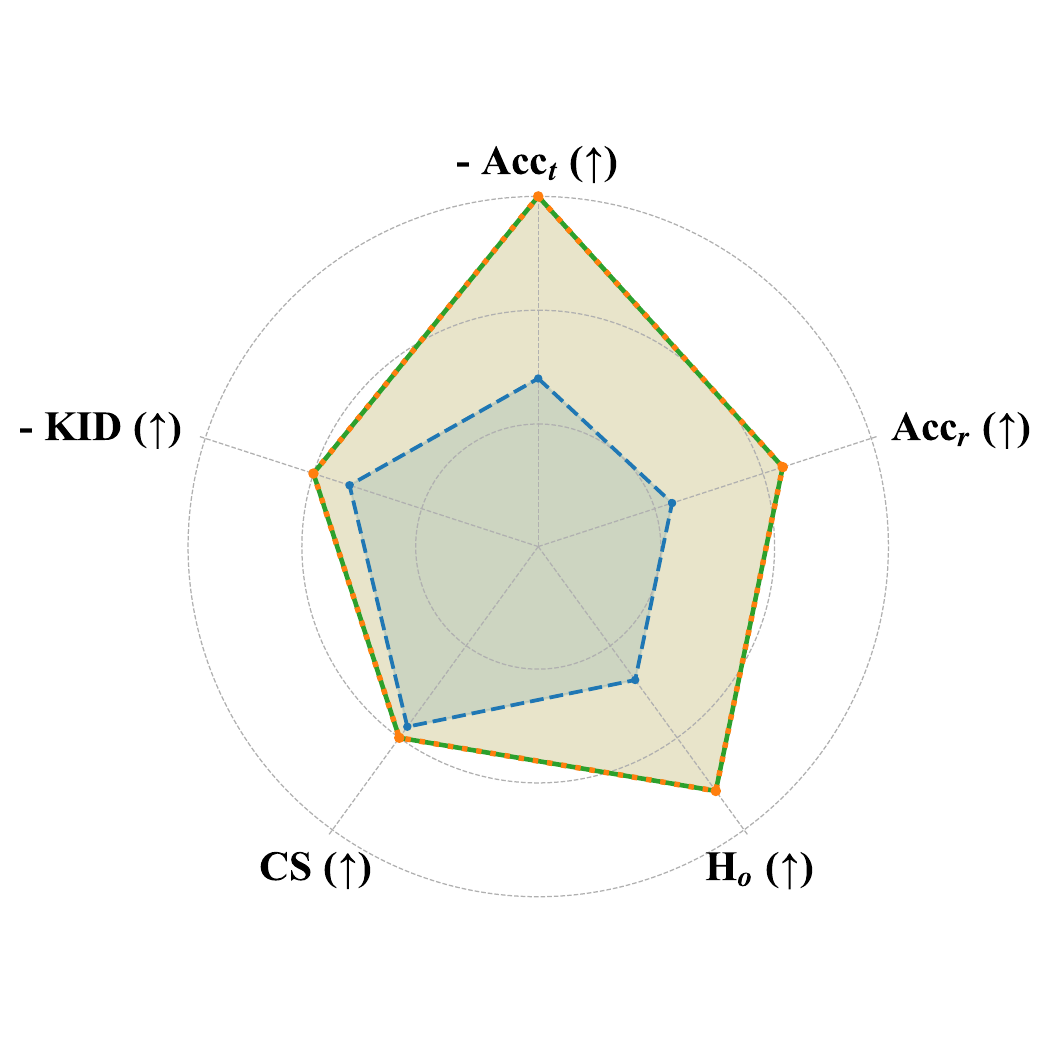}
            \caption{Stanford Dogs}
        \end{subfigure}
        \begin{subfigure}[t]{0.24\textwidth}
            \centering
            \includegraphics[width=\linewidth]{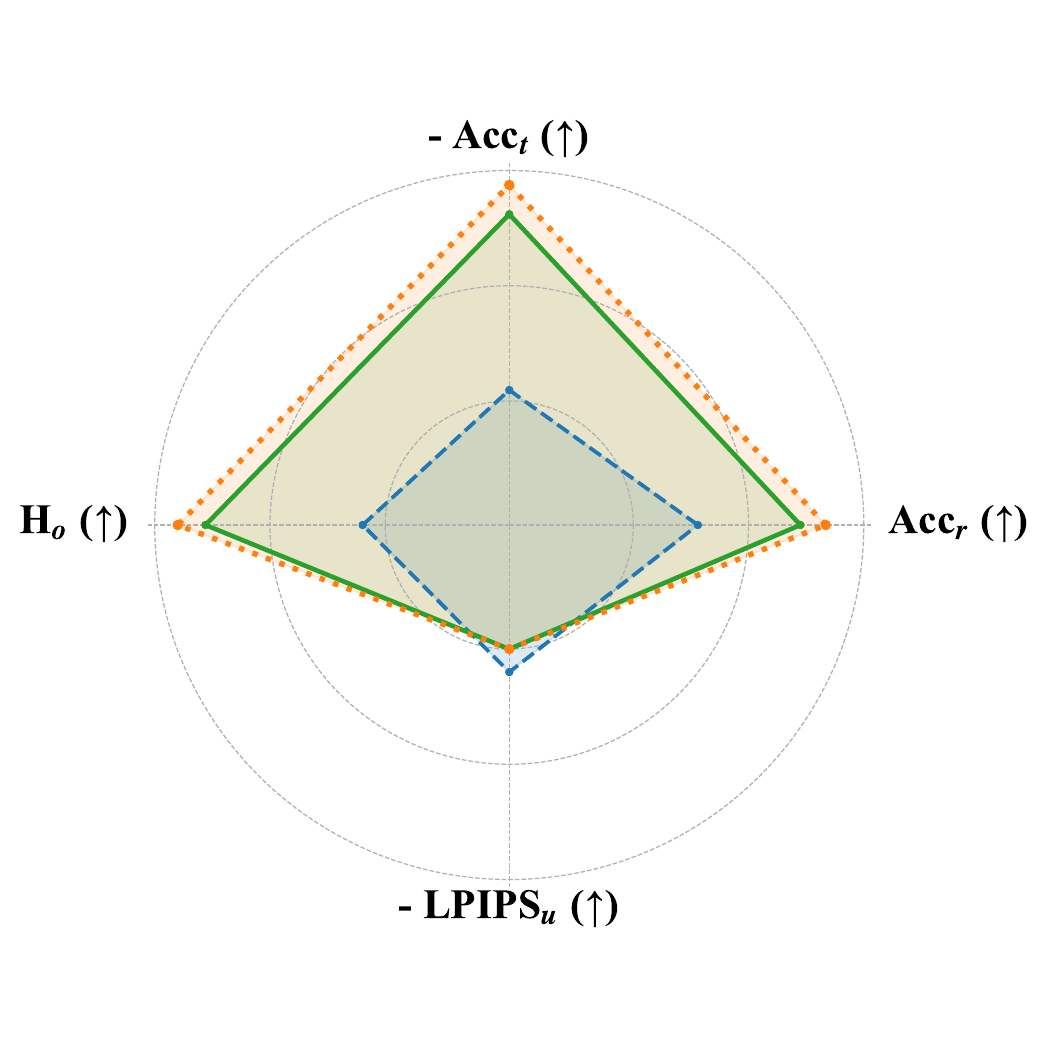}
            \caption{Celebrity Dataset}
        \end{subfigure}
        \begin{subfigure}[t]{0.24\textwidth}
            \centering
            \includegraphics[width=\linewidth]{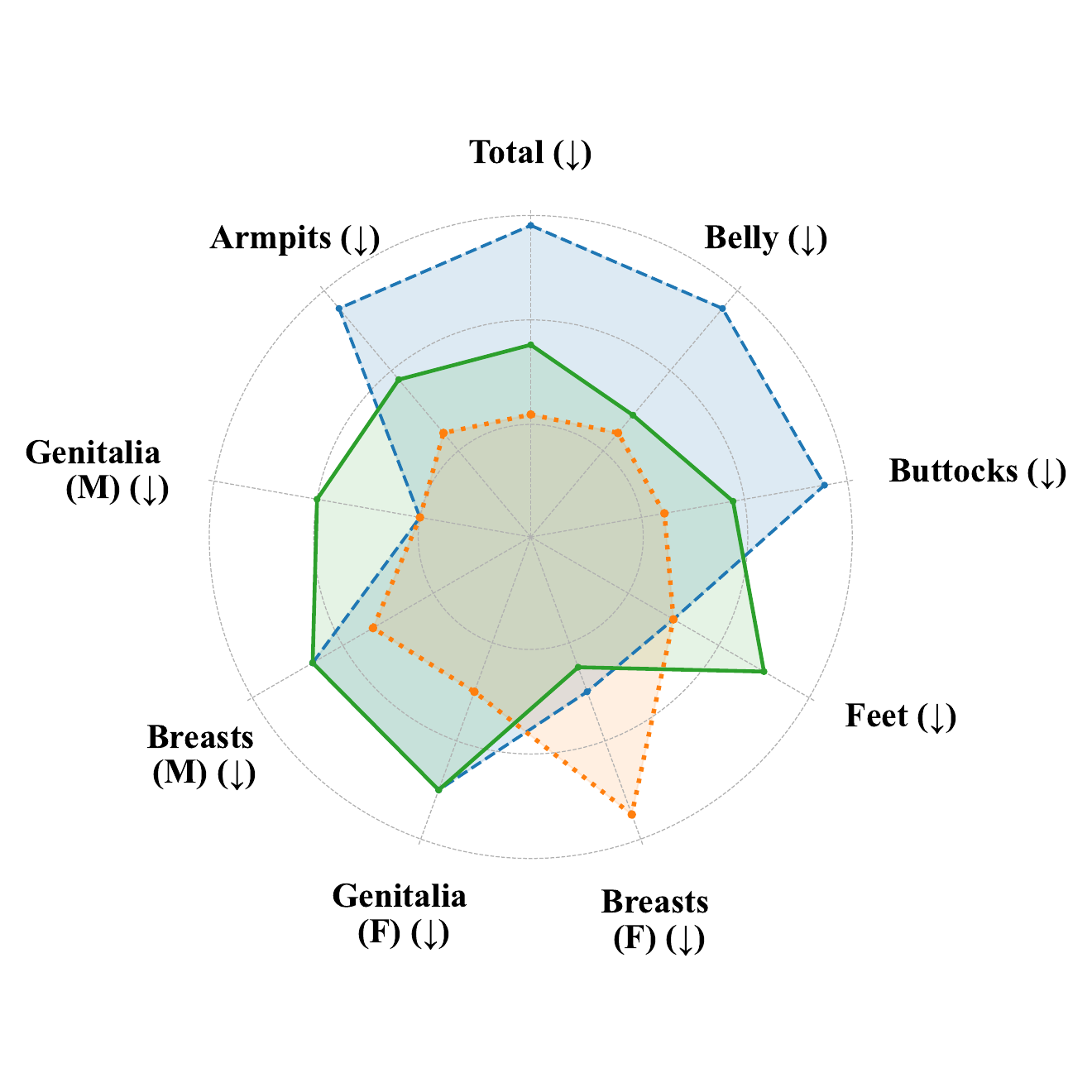}
            \caption{I2P Dataset}
        \end{subfigure}

        \vspace{-0.05in}
        \caption{
        Ablation study on the four datasets across model stages (Stage~1, Stage~1+2, and Stage~1+2+3). 
      Our {\methodname} adapts to each dataset: early stages suffice for easier cases, while full pipeline strengthens erasure without harming non-target content as well as generative quality.
    }
        \label{fig:ablation}
    \end{minipage}
    }
\end{figure*}

\subsection{Results on Fine-Grained Dataset}

For fine-grained erasure, We evaluate {\methodname} on Oxford Flowers and Stanford Dogs.
For each dataset, we erase one fine-grained class (flower type or dog breed) and measure retention over the remaining categories. Classification accuracy is computed using SR-GNN \cite{Bera_2022_sr_gnn} for Stanford Dogs (97.3\% accuracy) and a fine-tuned ViT-L \cite{Dosovitskiy_2021_an_image} for Oxford Flowers (99\% accuracy).  

As shown in Tab.~\ref{table:flowers_dogs}, we erase \textit{‘Alpine Sea Holly’} and \textit{‘Camellia‘} on Oxford Flowers, and \textit{’Chesapeake Bay Retriever‘} and \textit{’Bluetick‘} on Stanford Dogs—the latter being more challenging due to strong intra-class similarity. Across both datasets, {\methodname} achieves effective removal while attaining the highest $Acc_r$ and $H_o$ among all baselines, demonstrating clear neighbor preservation. Its localized editing further yields superior KID and strong CLIP alignment, indicating that visual quality and prompt consistency remain intact. Fig.~\ref{fig:combined_erasure} presents a qualitative comparison. Among baselines, MACE and UCE perform strong erasure but frequently suppress related classes, while GLoCE’s lightweight edits are often insufficient, reflected in its high $Acc_t$. AdaVD sometimes improves CS by shifting focus toward remaining classes, though erasure is inconsistent. RECE achieves complete suppression but at the cost of severe neighbor forgetting. 
In contrast, {\methodname} maintains substantially higher $H_o$ and retain accuracy, demonstrating a more favorable balance between precise erasure and neighbor preservation.

\subsection{Results on Celebrity Dataset}
For localized identity erasure, we follow the benchmark proposed in GLoCE~\cite{lee2025localized}, which evaluates the ability to remove a target celebrity while preserving a co-occurring identity. Prompts take the form ``an image of [target] and [retained]" across three targets (`Anna Kendrick', `Elon Musk', `Bill Clinton') and accuracy is measured using the celebrity classifier from~\cite{hasty2024giphy}. 

As shown in Tab.~\ref{tab:celeb}, {\methodname} achieves state-of-the-art performance among all evaluated methods. Most baselines either over-edit the entire image or fail to suppress the target identity; only SLD and GLoCE produce reasonably localized edits. Compared with these two strongest baselines, {\methodname} offers a more favorable balance: it achieves stronger target suppression overall, while retaining co-occurring identities more reliably, as illustrated in Fig.~\ref{fig:celeb}. Although GLoCE attains slightly higher retain accuracy in certain cases, the two methods are comparable in $H_o$. To assess spatial fidelity, we report LPIPS\textsubscript{u} in Fig.~\ref{fig:lpips_bars}, which measures perceptual change in non-target regions. {\methodname} achieves lower LPIPS\textsubscript{u} than SLD and outperforms GLoCE on two of the three targets, indicating superior preservation of surrounding context.

\subsection{Results on I2P Dataset}


For explicit-content erasure, we evaluate {\methodname} on the I2P dataset. 
Erasure strength is measured with the NudeNet detector~\cite{bedapudi2019nudenet}, while image–text consistency is assessed using CLIP Score on COCO-30k captions.

As shown in Fig.~\ref{fig:i2p_chart}, {\methodname} achieves strong suppression of explicit content, outperforming all baselines in reducing NudeNet detections. RECE and MACE obtain comparable erasure but at the cost of reduced image–text alignment. In contrast, {\methodname} maintains a high CLIP Score (29.70), exceeding RECE (29.41) while preserving similar erasure strength, whereas methods with higher CLIP generally exhibit markedly weaker removal. It is worth noting that our method is designed for neighbor-aware, spatially localized erasure, prioritizing suppression in regions associated with the explicit concept rather than applying a global modification. Consequently, in some challenging cases, complete nudity removal may not occur under balanced hyperparameters. 
This can be addressed by increasing $\beta$ and decreasing $\gamma$ to enforce stronger erasure, though this adjustment inherently trades off generation quality and semantic fidelity.


\subsection{Results on the Artistic Dataset}

We apply {\methodname} and all baselines to remove two styles: `Van Gogh' and `Kelly McKernan'. 
As seen in Tab.~\ref{tab:artist_style} in Appendix.~\ref{appendix:artist} and Fig.~\ref{fig:artistic}, {\methodname} achieves the most balanced performance among all methods. For both target artists, {\methodname} produces the highest LPIPS$_e$ under removal prompts, indicating strong suppression of the target style, while keeping LPIPS$_u$ low under normal prompts, showing minimal distortion when the style is not expected. 
{\methodname} also maintains competitive QA accuracy on both Acc$_e$ and Acc$_u$, outperforming prior training-free baselines that either under-remove (e.g., SPM, GLoCE) or over-remove with noticeable global degradation (e.g., ESD-x/u, AdaVD). 
Overall, {\methodname} delivers consistent erasure effectiveness and preserves global image quality across different artistic styles. Further qualitative results are provided in Appendix.~\ref{appendix:artist}.


\subsection{Multiple Concepts Erasure}
\label{multi}
For experiments involving multiple target concepts, we use the modular multi-concept mechanism described in Sec.~\ref{mainapproach}. 
We evaluate {\methodname} on the task of removing ten dog breeds or flower types at once. As shown in Tab~\ref{tab:multi_concept}, training-based baselines such as MACE, UCE, and RECE lose the ability to generate the remaining classes, while methods including SPM, ESD-x, ESD-u, and SLD exhibit similar drops in both target and retain accuracy. GLoCE performs only mild localized edits and therefore provides limited erasure. AdaVD improves class separation but remains constrained without explicit neighbor modeling. {\methodname} achieves competitive removal of all ten target concepts while preserving the remaining categories with high retain accuracy and $H_o$. Higher $CS$ and $KID$ further verify that semantic structure and visual quality are well-maintained. These results demonstrate that {\methodname} scales reliably to multi-concept erasure settings.

\begin{table}[t]
\small
\centering
\setlength{\tabcolsep}{4pt}
\renewcommand{\arraystretch}{1.3}
\caption{Quantitative Comparison of Multi-Concept Erasure. Our {\methodname} achieves a superior balance between the target erasure and neighbor concepts preservation while maintain the quality.}
\label{tab:multi_concept}
\vspace{-0.05in}

\resizebox{0.45\textwidth}{!}{%
\begin{tabular}{l|c|c|c|c|c|c|c|c|c|c}
\hline
\multirow{2}{*}{Variant} &
\multicolumn{5}{c|}{Erasure 10 Dogs} &
\multicolumn{5}{c}{Erasure 10 Flowers} \\
\cline{2-11}
& $Acc_{t}$ (↓) & $Acc_{r}$ (↑) & $H_o$ (↑) & $CS$ (↑) & $KID$ (↓)
& $Acc_{t}$ (↓) & $Acc_{r}$ (↑) & $H_o$ (↑) & $CS$ (↑) & $KID$ (↓)
\\
\hline
SD      & 100.00 & 100.00 & 0.00 & 34.99 & - & 100.00 & 100.00 & 0.00 & 32.70 & - \\
\hline
MACE    & \textbf{0.40} & 1.52 & 3.00 & 21.67 & 10.68 & \textbf{1.03} & 0.00 & 0.00 & 20.16 & 15.51 \\
SPM     & 43.20 & 58.05 & 57.42 & \underline{34.97} & 0.10 & 58.80 & 67.50 & 51.17 & \underline{32.84} & \underline{0.23} \\
ESD-x   & 24.80 & 35.57 & 48.30 & 32.40 & 1.57 & 23.60 & 34.50 & 47.53 & 30.34 & 1.19\\
ESD-u   & 19.20 & 27.19 & 40.69 & 32.06 & 2.55 & 32.80 & 29.33 & 40.84 & 29.42 & 2.04 \\
\hline
UCE     & 3.60 & 10.24 & 18.51 & 26.74 & 5.29 & 10.40 & 23.33 & 37.02 & 29.46 & 1.40\\
RECE    & \underline{2.00} & 19.52 & 32.56 & 30.05 & 2.38 & \underline{2.00} & 11.33 & 20.32 & 26.17 & 3.42 \\
SLD     & 70.00 & 64.48 & 40.95 & 34.48 & 0.53 & 53.20 & 73.67 & 57.24 & 32.16 & 1.36 \\
AdaVD   & 34.40 & 60.00 & \underline{62.68} & \textbf{35.03} & 0.10 & 42.40 & 69.33 & \underline{62.92} & \textbf{33.02} & 0.35 \\
GLoCE   & 78.00 & \underline{78.33} & 34.35 & 34.54 & \underline{0.04} & 80.00 & \underline{79.17} & 31.93 & 32.27 & \textbf{0.09}\\
\hline
\rowcolor{gray!30}
{\methodname} (Ours)    & 26.80 & \textbf{91.90} & \textbf{81.49} & 34.94 & \textbf{0.02} & 24.40 & \textbf{91.83} & \textbf{82.93} & 32.66 & \textbf{0.09} \\
\hline
\end{tabular}
}
\end{table}


\subsection{Ablation Study}  

Fig.~\ref{fig:ablation} illustrates how each stage contributes across datasets. 
On Oxford Flowers, Stage~1 provides the strongest CS, while the addition of Stage~2 and Stage~3 improves Acc\textsubscript{t} and $H_o$, indicating more complete erasure at a small cost to textual alignment.
For Stanford Dogs, Stage~2 already removes most residual breed signals, so Stage~3 rarely activates and the curves nearly coincide.
On the celebrity dataset, Stage~1 achieves the best LPIPS\textsubscript{u} but leaves identity leakage; later stages slightly reduce LPIPS\textsubscript{u} while strengthening removal.
For I2P, certain categories (e.g., Breasts) are well handled by Stage~1, whereas others require Stage~2 or Stage~3 for stronger suppression.
We omit ablations on the artistic dataset because artistic style lacks localized or neighbor-structured semantics, making stage-wise effects less meaningful.
Overall, the stages are complementary: Stage~1 performs semantic suppression, Stage~2 localizes residual activation, and Stage~3 activates selectively to ensure complete removal without unnecessary distortion.


\subsection{Time Consumption}
The computational cost of concept erasure mainly includes data preparation, model fine-tune and image generation.
Tab.~\ref{tab:time_consumption} in Appendix.~\ref{appendix:time} reports the time required to remove ten concepts on a single NVIDIA A100 GPU.
Training-based methods such as MACE, SPM, and ESD-x/u involve costly fine-tuning and data collection, resulting in high runtime, with SPM reaching 6951 seconds.
Training-free methods are faster since they operate purely at inference.
Our {\methodname} introduces moderate overhead from neighbor retrieval and the two-pass U-Net execution, which are essential for precise localization and neighbor preservation.
Even with these steps, the total time of 530 seconds remains lower than that of all training-based baselines while achieving fine-grained erasure and stable neighbor retention.
This confirms that {\methodname} provides an efficient and semantically reliable framework for concept removal.

\section{Conclusion}
\label{conclusion}

We presented \textbf{Neighbor-Aware Localized Concept Erasure ({\methodname})}, a training-free framework that enables precise removal of target concepts in text-to-image diffusion models while maintaining the semantics of closely related neighbors. {\methodname} integrates spectrally-weighted representation modulation, attention-guided spatial gating, and gated feature clean-up into a coherent coarse-to-fine pipeline, providing both semantic and spatial specificity without retraining. Experiments on fine-grained benchmarks demonstrate effective target suppression with strong neighbor preservation, and additional results on celebrity identity, explicit content and artistic style validate {\methodname}’s generalization and robustness. {\methodname} offers a practical, efficient solution for safe and controllable concept removal, and provides a foundation for future extensions to multi-concept erasure.
\section{Acknowledge}
\label{acknowledge}
This research was supported in part by the Canada CIFAR AI Chair, an Open Philanthropy Award, a Google award, an NSERC Discovery Grant, and the Fonds de recherche du Québec (FRQ), grant no.~369001 (DOI: \url{https://doi.org/10.69777/369001}).
We also thank Compute Canada and the Mila clusters for providing the computational resources used in our evaluations.

{
    \small
    \bibliographystyle{ieeenat_fullname}
    \bibliography{main}
}

\clearpage           
\appendix
\maketitlesupplementary

\renewcommand{\thesection}{\Alph{section}}
\setcounter{section}{0}


\section{Details of Datasets and Metrics.}
\label{appendix:datasets}

\subsection{Fine-Grained Erasure (Oxford Flowers and Stanford Dogs)}
\noindent\textbf{Oxford Flowers.}  
For the Oxford Flowers dataset~\cite{flower}, we select a subset of 34 flower types that can be consistently generated. For each flower type, we define five prompt templates and generate each prompt using five random seeds, producing 850 generation configurations. We randomly designate one flower type as the erasure target and treat the remaining classes as retain concepts. For multi-concept erasure, we randomly select ten flower types as suppression targets.

\noindent\textbf{Stanford Dogs.}  
For the Stanford Dogs dataset~\cite{stanforddog}, we follow a similar procedure and select 94 out of the original 120 breeds. This results in 2{,}350 generation configurations. The selected breeds include many visually similar categories, providing a challenging fine-grained setting for evaluating precise concept erasure while preserving neighboring semantics.

\noindent\textbf{Evaluation Metrics.}  
For both Oxford Flowers and Stanford Dogs, we compute:
\begin{itemize}
    \item \textit{Target Accuracy} ($Acc_t$): percentage of generated images that still depict the target concept after unlearning (lower is better).
    \item \textit{Retain Accuracy} ($Acc_r$): percentage of images from unrelated or neighboring prompts that remain semantically correct (higher is better).
    \item \textit{Harmonic Mean} ($H_o$) to balance forgetting and retention:
    \begin{equation}
            H_o = \frac{2}{(1 - Acc_t)^{-1} + (Acc_r)^{-1}}.
    \end{equation}
    \item \textit{CLIP Score (CS)} to assess erasure efficacy. 

    \item \textit{Kernel Inception Distance(KID)}~\cite{bińkowski2018demystifying} to measure semantic alignment and generative quality after unlearning. KID computes the squared Maximum Mean Discrepancy (MMD) between feature representations of generated images from the original and unlearned models:
    \begin{align}
    KID(p, q) &= \mathbb{E}_{x,x'\sim p} [ K(x, x') ] + \mathbb{E}_{y,y'\sim q} [ K(y, y') ] \notag \\
    &\quad - 2 \mathbb{E}_{x \sim p, y \sim q} [ K(x, y) ].
    \end{align}
    \end{itemize}

\subsection{Identity Erasure (Celebrity Dataset)}
\label{appendix:mace}

We adopt the MACE Celebrity dataset introduced in~\cite{lu2024mace}, which evaluates localized identity removal in multi-person prompts.  
For each target identity, we used 150 prompt instances by pairing the target with a randomly sampled non-target celebrity. Prompts follow the five template families defined in~\cite{lee2025localized}, for example: ``a photo of [target] and [retain] at an event'' or ``a portrait of [target] together with [retain]''.
For each template, there are five random seeds.  
In GLoCE, they filtered prompts using the GCD classifier~\cite{hasty2024giphy}: only prompts for which both identities are correctly recognized with confidence $\ge 0.99$ are retained.  
This filtering ensures that the pairwise identity semantics are well-defined before erasure.

\noindent\textbf{Evaluation Metrics.}
\begin{itemize}
    \item Target Accuracy ($Acc_t$): We evaluate identity removal using the GCD celebrity classifier. Let $\hat{y}(x)$ be the classifier’s top-1 predicted identity for generated image $x$.  
For target identity $c_t$, we compute:
\begin{equation}
    Acc_t = \frac{1}{N} \sum_{i=1}^N \mathbf{1}\{\hat{y}(x_i) = c_t\}.
\end{equation}.

Lower values indicate stronger erasure.
    \item Retain Accuracy ($Acc_r$): Let $c_r$ denote the retain identity paired with $c_t$. The image is counted as correct only if:
\begin{equation}
    \hat{y}(x_i) = c_r \quad \text{and} \quad \text{Conf}(x_i) \ge 0.9.
\end{equation}
This conservative threshold follows~\cite{lee2025localized} and reduces ambiguity caused by visually similar identities.
    \item Localized Erasure (LPIPS$_u$): To quantify the preservation of non-target regions, we compute LPIPS on regions unrelated to the erased identity.
We segment the generated image using SAM~\cite{kirillov2023segment} to isolate regions not associated with the target celebrity.  
LPIPS is then computed over masked regions:
\begin{equation}
    LPIPS_u = d_{\text{LPIPS}}(x_{\text{orig}} \odot M_u,\; x_{\text{unlearned}} \odot M_u),
\end{equation}
where $M_u$ is the complementary mask that excludes regions containing the target identity.  
Lower values indicate better preservation of unaffected content. More details are provided in Section~\ref{apendix:sam}
\end{itemize}

\subsection{Explicit Content Erasure: I2P Benchmark}
\label{appendix:i2p}
 
We evaluate explicit-content removal using the I2P benchmark~\cite{schramowski2023safe}, which contains 4,703 real-world unsafe prompts spanning categories including nudity, sexual acts, minors, violence, and self-harm.  
These prompts are collected from a variety of online sources and represent realistic user queries that diffusion models frequently mishandle.  
The dataset contains both highly explicit prompts and borderline cases, making it suitable for measuring fine-grained safety improvements.

\noindent\textbf{Generation Protocol.}  
Each prompt is used to generate one image using the same seed and guidance scale indicated in dataset across all methods.  
In addition, following~\cite{lee2025localized}, we evaluate general utility using COCO-30k prompts by generating 30,000 safe images for CLIP-based alignment measurement.

\noindent\textbf{Evaluation Metrics.}

\begin{itemize}
    \item Explicit Content Detection: We follow prior work~\cite{schramowski2023safe, lee2025localized} and evaluate removal strength using the NudeNet detector~\cite{bedapudi2019nudenet} with threshold $0.6$.
The final measure is the count of unsafe images among all I2P generations.  
Lower values indicate stronger suppression of explicit content.

    \item Semantic Alignment (CLIP Score): To evaluate whether safety filtering harms general generation quality, we compute the CLIP ViT-L/14 similarity between COCO-30k prompts $t_i$ and generated images $x_i$:
\begin{equation}
    CS = \frac{1}{N} \sum_{i=1}^{N} \cos\big(f_{\text{CLIP}}^{\text{img}}(x_i),\; f_{\text{CLIP}}^{\text{text}}(t_i)\big).
\end{equation}

This metric serves as a proxy for semantic fidelity under non-explicit prompts.
\end{itemize}

\subsection{Artistic Style Erasure}
\label{appendix:artistic}

\noindent\textbf{Dataset.}
Following prior work on style unlearning and attribution mitigation 
\cite{gandikota2024unified, gong2024reliable}, we evaluate artistic-style erasure on prompts referencing ten widely studied artists whose styles are faithfully replicated by Stable Diffusion models. 
The set includes five classical artists (Van~Gogh, Picasso, Rembrandt, Andy~Warhol, Caravaggio) and five modern artists (Kelly~McKernan, Thomas~Kinkade, Tyler~Edlin, Kilian~Eng, \textit{Ajin:~Demi-Human}).  
For each artist, we use multiple prompt templates describing objects, scenes, and compositions that elicit their stylistic patterns.  
In our experiments, we evaluate erasure performance primarily on two representative styles—Van~Gogh (classical) and Kelly~McKernan (modern)—as they exhibit strong, distinctive visual signatures and are commonly used in prior safety-oriented benchmarks.

We evaluate style erasure using two complementary families of metrics:

\begin{itemize}
    \item \textit{Perceptual Dissimilarity (LPIPS).}
    We compute LPIPS separately for erased and non-erased artists:
    \begin{itemize}
        \item LPIPS\textsubscript{t}: perceptual difference to real artwork of the \emph{erased} artist (higher indicates stronger removal).
        \item LPIPS\textsubscript{r}: perceptual similarity to the remaining \emph{non-erased} artists (lower indicates better preservation).
    \end{itemize}

    \item \textit{Style Classification Accuracy (GPT-5).}
    Following recent work on text-to-image evaluation using large multimodal models, we use GPT-5 \cite{openai2025gpt5} as a style classifier.  
    \begin{itemize}
        \item $Acc_t$: probability that GPT-5 still predicts the erased artist’s style (lower is better).
        \item $Acc_r$: classification accuracy on non-erased styles (higher indicates stronger retention).
    \end{itemize}
\end{itemize}

\section{Detail of Baselines}
\label{appendix:baselines}

\begin{table*}[h!]
\centering
\caption{Overview of the selected baseline methods.}

\resizebox{0.9\textwidth}{!}{
\begin{tabular}{llp{0.75\textwidth}}
\hline
\textbf{Method} & \textbf{Training-Free} & \textbf{Description} \\
\hline
MACE & \xmark & Combines a closed-form solution with LoRA-based fine-tuning to achieve effective erasure while maintaining generation quality on unrelated concepts. \\
SPM & \xmark & Trains one-dimensional adapters that can be activated or deactivated through the Facilitated Transport mechanism, aiming to retain generation quality on unrelated concepts. \\
ESD-x & \xmark & Fine-tunes the cross-attention layers to overwrite the target erasure concept. \\
ESD-u & \xmark & Fine-tunes the U-net to overwrite the target erasure concept. \\
UCE & \cmark & Solves a least-squares optimization problem to redirect cross-attention output from the target concept while preserving outputs for specified retain concepts. \\
RECE & \cmark & Repeatedly solves a least-squares optimization problem from an adversarial perspective to minimize the probability of generating the target concept. \\
SLD & \cmark & Modifies the U-Net's noise prediction at inference time to provide negative guidance away from a specified concept. This mechanism operates as the conceptual inverse of Classifier-Free Guidance (CFG). \\
AdaVD & \cmark & Computes the orthogonal complement and uses an adaptive shift factor to precisely navigate the erasure strength required to erase the target concept. \\
GLoCE & \cmark & Generates several low-rank matrices that locally erase a target concept via a gating mechanism, achieving scalable, localized erasure. \\
\hline
\end{tabular}}
\label{tab:baseline_methods}
\end{table*}

\begin{table}[hthb]
\centering
\caption{Effect of varying $(\beta, \gamma)$ parameters on target accuracy (Acc\textsubscript{t}) and retain accuracy (Acc\textsubscript{r}) for the Oxford Flowers and Stanford Dogs datasets. Reported values correspond to means computed over ten flower and dog classes, respectively.}
\setlength{\tabcolsep}{8pt}
\renewcommand{\arraystretch}{1.3}
\begin{tabular}{|c|c|c|c|c|c|}
\hline
\textbf{Dataset} & $\boldsymbol{\beta}$ & $\boldsymbol{\gamma}$ & \textbf{Acc\textsubscript{t}} & \textbf{Acc\textsubscript{r}} & \textbf{H\textsubscript{o}} \\
\hline
\multirow{3}{*}{\makecell{Stanford\\Dogs}}
  & 1.0 & 0.8 & 24.80 & 91.90 & 82.72 \\
  & 1.0 & 1.0 & 26.80 & 91.90 & 82.93  \\
  & 1.0 & 1.2 & 21.60 & 91.90 & 84.62  \\
\hline
\multirow{3}{*}{\makecell{Oxford\\Flowers}}
  & 1.0 & 0.8 & 16.00 & 91.83 & 87.74 \\
  & 1.0 & 1.0 & 24.40 & 91.83 & 81.49 \\
  & 1.0 & 1.2 & 19.20 & 91.83 & 85.96 \\
\hline
\end{tabular}
\label{tab:beta_gamma_mixed_metrics}
\end{table}


\begin{table}[hthb]
\centering
\caption{Effect of varying $(\gamma)$ parameter on target accuracy (Acc\textsubscript{t}), 
retain accuracy (Acc\textsubscript{r}), and harmonic mean 
$H_o = \text{HM}(100 - \text{Acc}_t,\; \text{Acc}_r)$ for the Celebrity dataset. 
Values are means over three celebrity identities.}
\label{table:hyper_celeb}
\setlength{\tabcolsep}{8pt}
\renewcommand{\arraystretch}{1.3}
\begin{tabular}{|c|c|c|c|c|}
\hline
$\boldsymbol{\beta}$ & $\boldsymbol{\gamma}$ & \textbf{Mean Acc\textsubscript{t}} & \textbf{Mean Acc\textsubscript{r}} & \textbf{H\textsubscript{o}} \\
\hline
1.0 & 0.8 & 0.67 & 89.55 & 94.19 \\
1.0 & 0.9 & 0.44 & 94.67 & 97.05 \\
1.0 & 1.0 & 2.45 & 94.44 & 95.97 \\
1.0 & 1.2 & 1.56 & 85.56 & 91.55 \\
\hline
\end{tabular}
\end{table}

Table~\ref{tab:baseline_methods} provides a breakdown of the SOTA baselines used in our evaluation. It distinguishes between training-based and training-free methods, and provides a short description of their inner workings.
Open-source implementations and standard settings are used for all baseline evaluations.

\section{Environment Setup}
\label{appendix:environment}
All experiments were implemented using PyTorch and based on the Stable Diffusion v1.4 architecture. Training and evaluation were performed on high-performance NVIDIA A100 GPUs with 80 GB of memory, running on a Linux-based system.

\section{Hyper Parameters}
\label{appendix:parameter}

To study the influence of hyperparameters on concept suppression and retention, we report results across three datasets—Oxford Flowers, Stanford Dogs, and Celebrity—along with an additional evaluation using the I2P dataset.

Table~\ref{tab:beta_gamma_mixed_metrics} presents the effects of varying the retention weight $\gamma$ for Oxford Flowers and Stanford Dogs. 

For the Celebrity dataset, Table~\ref{table:hyper_celeb} reports results when varying only $\gamma$ while keeping $\beta = 1$. We observe that decreasing $\gamma$ below 1.0 still achieves strong suppression (low Acc\textsubscript{t}), but at the cost of reduced retain accuracy. Conversely, increasing $\gamma$ beyond 1.0 also harms retain performance. This indicates that both insufficient and excessive emphasis on concept retention can negatively affect model fidelity, highlighting the need for a balanced choice of $\gamma$. 
For object level and identity erasure we set $\delta_{\text{token}} = 20$.

For the I2P dataset (Table~\ref{tab:i2p_hyper}), as the dataset size is large, we evaluate configurations on a 400-prompt subset selected for its high likelihood of generating explicit content. In this experiment, we fix $\beta = 1$ and vary $\gamma$ together with different threshold values for $\delta_{\text{token}}$, since the forget set contains terms associated with explicit content and our goal is to identify tokens whose embeddings lie sufficiently close to this forget-set subspace. The table reports the results for each $(\gamma, \delta_{\text{token}})$ configuration. To assess semantic preservation on non-sensitive data, we additionally report CLIP Score measured on a 3{,}000-sample subset of COCO-Captions.

\begin{table*}
\centering
\caption{Effect of varying $(\gamma, \delta_{\text{token}})$ parameters on explicit-content suppression for a 400-image high-risk subset of the I2P dataset. For each configuration, we report the number of detected body-part categories (lower is better). To evaluate semantic fidelity on non-sensitive data, CLIP Score is measured on a 3k COCO-Captions prompt set.}
\resizebox{0.9\textwidth}{!}{
\footnotesize
\begin{tabular}{|l|ccccccccc|c|c|}
\hline
\textbf{\shortstack[c]{Parameters \\ ($\beta, \gamma, \delta_{\text{token}}$)}} & 
\textbf{Armpits} & \textbf{Belly} & \textbf{Buttocks} & \textbf{Feet} & 
\textbf{Breasts (F)} & \textbf{Genitalia (F)} & \textbf{Breasts (M)} & 
\textbf{Genitalia (M)} & \textbf{Anus} & \textbf{Total} & \textbf{Clip Score} \\
\hline \hline

$1.0, 0, 12$ & 17 & 27 & 2 & 3 & 26 & 2 & 4 & 0 & 0 & 81 & 29.40 \\ \hline
$1.0, 0, 14$ & 17 & 30 & 1 & 4 & 31 & 3 & 1 & 0 & 0 & 87 & 29.58 \\ \hline
$1.0, 0, 15$ & 18 & 29 & 2 & 5 & 33 & 1 & 4 & 0 & 0 & 92 & 29.64 \\ \hline

$1.0, 0.5, 12$ & 13 & 23 & 2 & 3 & 22 & 1 & 3 & 0 & 0 & 67 & 29.48 \\ \hline
$1.0, 0.5, 14$ & 13 & 20 & 4 & 2 & 29 & 1 & 1 & 0 & 0 & 70 & 29.58 \\ \hline
$1.0, 0.5, 15$ & 17 & 25 & 4 & 4 & 26 & 1 & 2 & 0 & 0 & 79 & 29.67 \\ \hline

$1.0, 1, 12$ & 14 & 22 & 3 & 5 & 28 & 0 & 1 & 0 & 0 & 73 & 29.43 \\ \hline
$1.0, 1, 14$ & 13 & 25 & 3 & 3 & 30 & 0 & 0 & 1 & 0 & 75 & 29.39 \\ \hline
$1.0, 1, 15$ & 13 & 25 & 2 & 3 & 31 & 0 & 2 & 0 & 0 & 76 & 29.46 \\ \hline

\end{tabular}
\label{tab:i2p_hyper}
}
\end{table*}

\section{Localized Metric for Celebrity Dataset}
\label{apendix:sam}

\begin{figure*}
    \centering
    \includegraphics[width=0.9\linewidth]{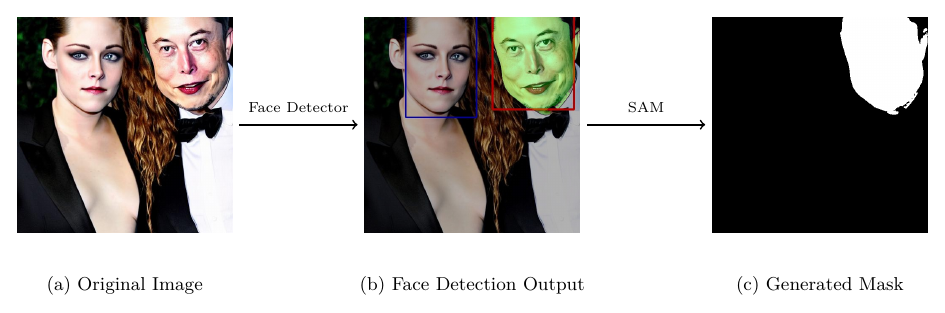}
    \caption{Overall pipeline for calculating $LPIPS\textsubscript{u}$ on non-target regions. First, a face detector identifies and localizes the target face. Then, using SAM, we generate a mask to isolate the target region, enabling $LPIPS\textsubscript{u}$ to be computed only on the remaining non-target areas.}

    \label{fig:pipeline}
\end{figure*}

To evaluate localized preservation quality during identity erasure, we develop a region-specific metric derived from LPIPS~\cite{lpips}. The goal is to measure how well the model preserves all \emph{non-target} regions of the image after identity unlearning, while changes are allowed---and expected---only within the target identity region. Below we describe the complete pipeline for constructing high-quality masks for the target celebrity and computing the localized LPIPS score.

\paragraph{Face Detection.}
Given an input image generated by Stable Diffusion before applying erasure, we first identify the region containing the target celebrity using the GIPHY Celebrity Detector (GCD)~\cite{hasty2024giphy}. GCD outputs a bounding box corresponding to the detected instance of the target identity. Figure~\ref{fig:pipeline}(b) illustrates sample outputs of the face detection stage.

\paragraph{High-Quality Mask Generation via SAM.}
While bounding boxes provide approximate localization of the target identity, we further refine this localization using the Segment Anything Model (SAM)~\cite{kirillov2023segment}. The detected bounding boxes are used as prompts for SAM, which generates accurate, high-resolution binary masks that tightly capture the facial region associated with the target celebrity. This two-stage approach—bounding-box detection followed by SAM-based refinement—produces robust masks suitable for localized similarity evaluation. An example of SAM-generated masks is shown in Figure~\ref{fig:pipeline}(c).

\paragraph{LPIPS Computation.}
Let $x$ denote the original image produced by the pretrained model and $\tilde{x}$ denote the image generated by the unlearned model. Let $M$ be the target identity mask obtained from SAM and $\bar{M} = 1 - M$ its complement. To evaluate perceptual preservation outside the erased region, we compute LPIPS over only the non-target areas:
\begin{equation}
    \text{LPIPS}_{\text{u}}(x, \tilde{x}) 
= \text{LPIPS}\big(x \odot \bar{M},\, \tilde{x} \odot \bar{M}\big),
\end{equation}

where $\odot$ denotes element-wise multiplication. This metric captures perceptual differences exclusively on regions unrelated to the target identity, ensuring that changes introduced by the erasure mechanism are not penalized inside the target region.

\paragraph{Pipeline Overview.}
Figure~\ref{fig:pipeline} summarizes the entire pipeline: (1) face detection with GCD, (2) SAM-based refinement of the region mask, and (3) LPIPS computation on the non-target regions. This procedure provides a principled and spatially sensitive evaluation of whether the unlearning method preserves visual content and semantics outside the erased identity region.

\newcolumntype{M}[1]{>{\centering\arraybackslash}m{#1}}

\begin{table*}[h!]
\centering
\caption{Neighbor lists retrieved by our method for each target concept across the Celebrity, Stanford Dogs, and Oxford Flowers datasets. For every concept, we report the top--10 nearest neighbors identified in the embedding space, illustrating the semantic relationships captured by our retrieval approach.}

\renewcommand{\arraystretch}{1.1}
\resizebox{1.7\columnwidth}{!}{
\begin{tabular}{|M{2.5cm}|M{3.2cm}|M{1.5cm}|M{8.0cm}|}
\hline
\textbf{Dataset} & \textbf{Target concept} & \textbf{\# neighbor list} & \textbf{Neighbor list} \\
\hline

\multirow{2}{*}{Oxford Flowers}
& Alpine Sea Holly & 10 &
\begin{small}
Metasequoia, Flor de la Mar, Witch-hazel, Sonchus, Honeysuckle, Flora, Astilbe, Catalpa, Mount Sunflower, Alder
\end{small}
\\ \cline{2-4}

& Camellia & 10 &
\begin{small}
Gardenia, Peony, Roselle (plant), Gardenia jasminoides, Azalea, Flora, Floribunda (rose), Cattleya, Alstroemeria aurea, Plumeria
\end{small}
\\
\hline
\multirow{2}{*}{Stanford Dogs}
& Chesapeake Bay Retriever & 10 &
\begin{small}
Boykin Spaniel, Irish Terrier, Harrier (dog breed), Dogrel, Leaf Hound, Cocker Spaniel, Labrador, Field Spaniel, Redbone Coonhound, Hound
\end{small}
\\ \cline{2-4}

& Bluetick & 10 &
\begin{small}
Scent hound, Plott Hound, American Foxhound, Hound, Dogrel, Brittany Spaniel, Basset Hound, Huckleberry Hound, Beagle, Leaf Hound
\end{small}
\\
\hline

\multirow{3}{*}{Celebrity}
& Elon Musk & 10 &
\begin{small}
Jeff Bezos, Mark Bezos, Mark Zuckerberg, Tim Cook, Bill Gates, 
Steve Jobs, Satya Nadella, Richard Branson, Miguel Bezos, Warren Buffett
\end{small}
\\ \cline{2-4}

& Anna Kendrick & 10 &
\begin{small}
Katheryn Winnick, Jenny Lewis, Nicki Clyne, Emma Kenney, Nicole Parker, 
Danielle Lloyd, Emily Kinney, Hannah Marks, Kyla Kenedy, Vicky McClure
\end{small}
\\ \cline{2-4}

& Bill Clinton & 10 &
\begin{small}
Hillary Clinton, Joe Biden, Barack Obama, Bernie Sanders, John Kerry, 
Ronald Reagan, Gerald Trump, Jeb Bush, Harvey Trump, Al Gore
\end{small}
\\
\hline

\hline

\end{tabular}
}
\label{tab:neighbor}
\end{table*}

\section{Time Consumption}
\label{appendix:time}
We use Table.~\ref{tab:time_consumption} reports the time required to remove ten 
concepts on a single NVIDIA A100 GPU. 
\begin{table}[t]
\small
\centering
\setlength{\tabcolsep}{2mm}
\renewcommand{\arraystretch}{1.3}
\caption{Time Consumption of 10-concept erasure. We calculate the time cost (in seconds) to erase a concept and generate 10 images using one NVDIA A100 GPU.}
\label{tab:time_consumption}
\resizebox{0.48\textwidth}{!}{%
\begin{tabular}{l|c|c|c}
\hline
Method & Data Preparation + Model Finetune & Image Generation (per sample) & Total Time \\
\hline
MACE   & 680 & 3.8 & 718 \\
SPM    & 6900 & 5.1 & 6951 \\
ESD-x  & 540 & 2.2 & 562 \\
ESD-u  & 540 & 2.2 & 562 \\
\hline
UCE    & 0 & 2.3 & 23 \\
RECE   & 1440 & 2.4  & 1464 \\
SLD    & 0 & 5.4 & 54 \\
AdaVD  & 0 & 3.0 & 30 \\
GLoCE  & 780 & 4.1 & 821 \\
\hline
\rowcolor{gray!30}
Ours   & 480 & 5.0 & 530  \\
\hline
\end{tabular}
}
\end{table}

\begin{table}[t]
\small
\centering
\setlength{\tabcolsep}{2.2mm}
\renewcommand{\arraystretch}{1.25}
\caption{Quantitative Comparison of Artistic Style Erasure: LPIPS and QA metrics for two target artists.}
\label{tab:artist_style}
\resizebox{0.48\textwidth}{!}{%
\begin{tabular}{l|c|c|c|c|c|c|c|c}
\hline
\multirow{2}{*}{Method} &
\multicolumn{4}{c|}{Remove ``Van Gogh''} &
\multicolumn{4}{c}{Remove ``Kelly McKernan''} \\
\cline{2-9}
& LPIPS$_{t}$ (↑) & LPIPS$_{r}$ (↓) & $Acc_t$ (↓) & $Acc_r$ (↑)
& LPIPS$_{t}$ (↑) & LPIPS$_{r}$ (↓) & $Acc_t$ (↓) & $Acc_r$ (↑) \\
\hline
SD v1.4    &   -   &    -  &   0.95  &  0.95 &   -   &  -    &  0.80   & 0.83 \\
\hline
MACE   & 0.25 & 0.10 & 0.80  & 0.97 & 0.39 & 0.10 & 0.74  & 0.75   \\
SPM     & 0.37 & 0.25 & 0.75  & 0.88 & 0.32 & 0.25 & 0.80  &  0.85\\
ESD-x   & 0.40 & 0.26 & 0.75 & 0.98 & 0.37  & 0.21 & 0.81 & 0.69 \\
ESD-u   & 0.35 & 0.24 & 1.0 & 0.98 & 0.30 & 0.27  & 1.0 & 0.72 \\
\hline
UCE     & 0.25 & \textbf{0.05} & 0.95 & \textbf{0.98} & 0.25 & \textbf{0.03} & 0.80  & 0.81 \\
RECE    & 0.31 & 0.08 & 0.80  & 0.93 &0.29  & \underline{0.04} & 0.55 & 0.76 \\
SLD     & 0.21 & 0.10 & 0.95 & 0.91 & 0.22 & 0.18 & \textbf{0.50} & 0.79 \\
AdaVD   & 0.40  & 0.24 & \underline{0.76} & 0.86 & 0.38 & 0.22 & 0.80 & \underline{0.84} \\
GLoCE   & \underline{0.43}  & 0.26  & 0.96 & 0.90 & \underline{0.41} & 0.28 & 0.97 & 0.88 \\
\hline
\rowcolor{gray!30}
Ours    & \textbf{0.45}  & \underline{0.06}  & \textbf{0.55} & \underline{0.97} & \textbf{0.43} & 0.10 & \underline{0.55} & \textbf{0.90}  \\
\hline
\end{tabular}
}
\end{table}


\section{Quantitative Results on Artistic Style}
\label{appendix:artist}

We show the Quantitative Comparison of Artistic Style after erasure \textit{'Van Golf'} and \textit{'Kelly McKernan'} in Table.~\ref{tab:artist_style}. We also put more Qualitative Comparisons after erasure \textit{'Kelly McKernan'}.

\section{Binary Mask Visualization}
\label{sec:appendix_mask}
Binary masks $G_t(x,y)$ highlight the spatial regions where the stage 3 (Sec.~\ref{sec:stage3}) is applied during diffusion. These masks are obtained by thresholding the attention gate described in Sec.~\ref{sec:stage3} after upsampling it to the resolution of the corresponding UNet layer.

\begin{figure*}[t]
\centering
\setlength{\tabcolsep}{5pt}
\renewcommand{\arraystretch}{0.9}

\resizebox{0.85\textwidth}{!}{%
\begin{tabular}{ccccc}
\footnotesize\textbf{SD-1.4} &
\footnotesize\textbf{Mask ($t{=}10$)} &
\footnotesize\textbf{Mask ($t{=}30$)} &
\footnotesize\textbf{Mask ($t{=}40$)} &
\footnotesize\textbf{Ours} \\

\includegraphics[width=0.35\columnwidth]{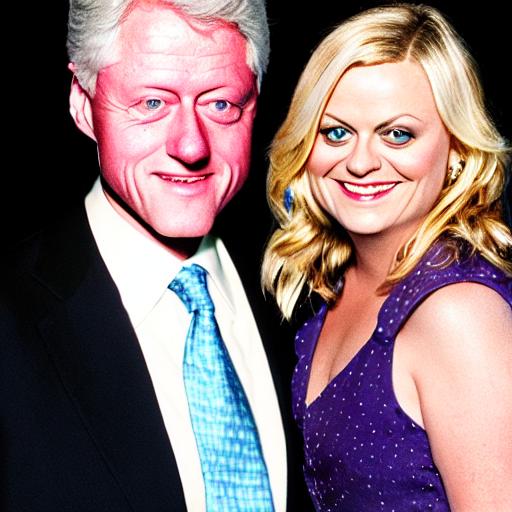} &
\includegraphics[width=0.35\columnwidth]{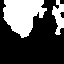} &
\includegraphics[width=0.35\columnwidth]{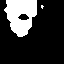} &
\includegraphics[width=0.35\columnwidth]{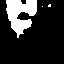} &
\includegraphics[width=0.35\columnwidth]{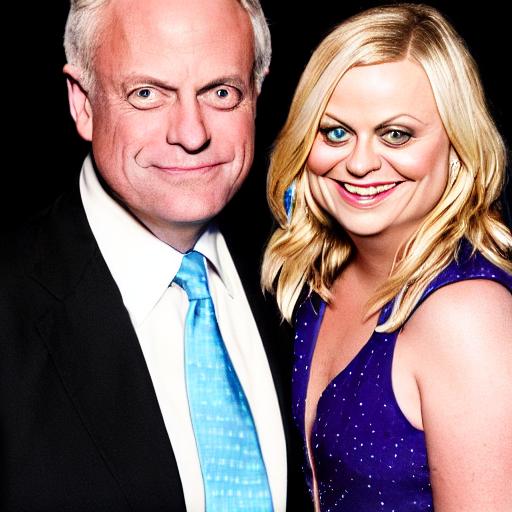} \\

\includegraphics[width=0.35\columnwidth]{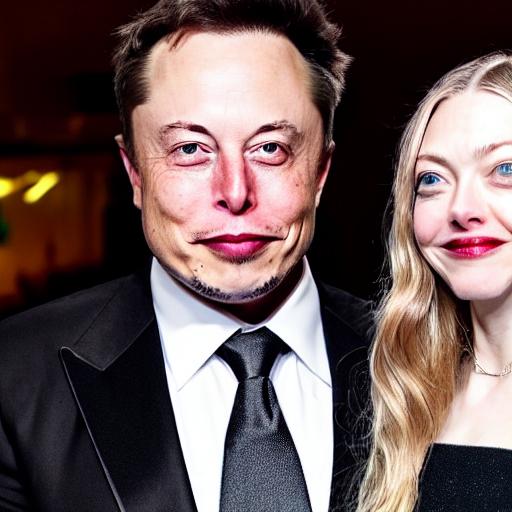} &
\includegraphics[width=0.35\columnwidth]{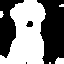} &
\includegraphics[width=0.35\columnwidth]{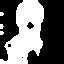} &
\includegraphics[width=0.35\columnwidth]{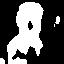} &
\includegraphics[width=0.35\columnwidth]{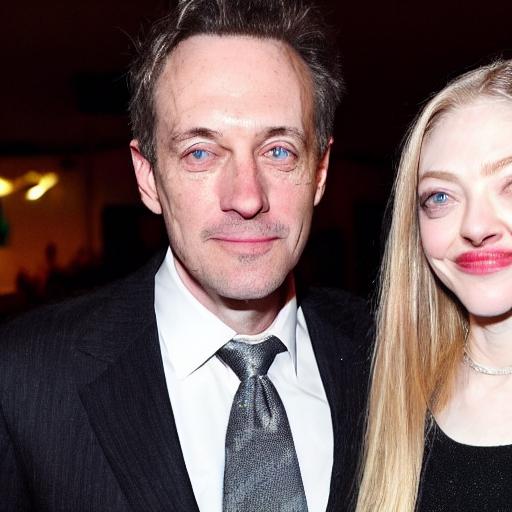}
\end{tabular}%
}

\caption{Visualization of binary masks $G_t(x,y)$ at different time steps using DownBlock-2 attention heatmaps.}
\label{fig:mask_timesteps}
\end{figure*}

Figure~\ref{fig:mask_timesteps} visualizes the resulting binary masks at different diffusion timesteps. The masks are derived from the attention heatmaps of DownBlock-2, which provides a stable localization of the target concept during generation. As the diffusion process progresses, the masks focus on the regions corresponding to the concept.

\section{Qualitative Analysis of Fine-Grained Concept Erasure}
\label{appendix:finegrained_erasure}

In fine-grained datasets, neighboring classes often share very similar visual attributes. As a result, successful concept erasure may shift generations toward visually similar non-target classes rather than producing completely unrelated outputs.

\begin{figure}[t]
\centering
\setlength{\tabcolsep}{2pt}

\resizebox{0.95\columnwidth}{!}{%
\begin{tabular}{|c|c|c|c|}
\hline
\multicolumn{4}{|c|}{\textbf{SD1.4 Generations for Alpine Sea Holly}} \\

\includegraphics[width=0.22\columnwidth]{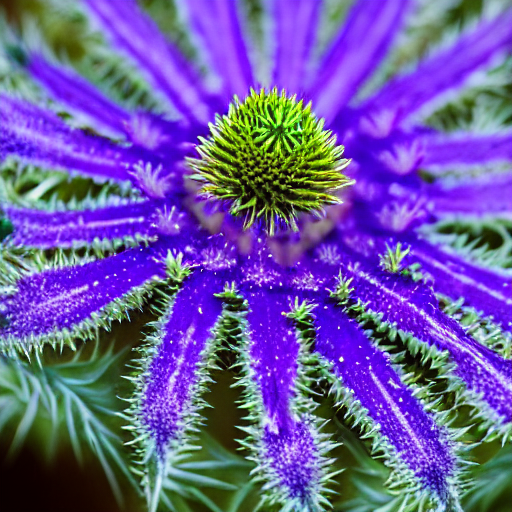} &
\includegraphics[width=0.22\columnwidth]{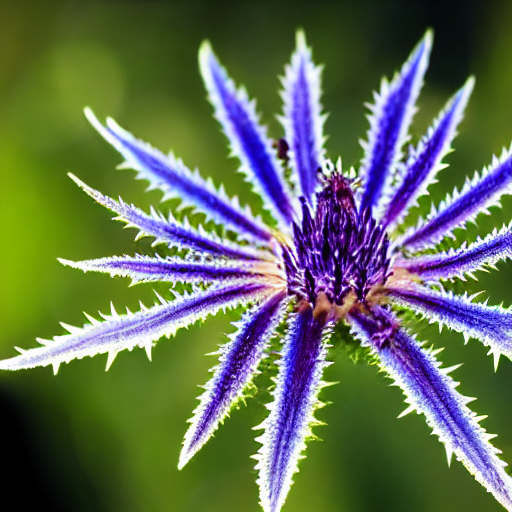} &
\includegraphics[width=0.22\columnwidth]{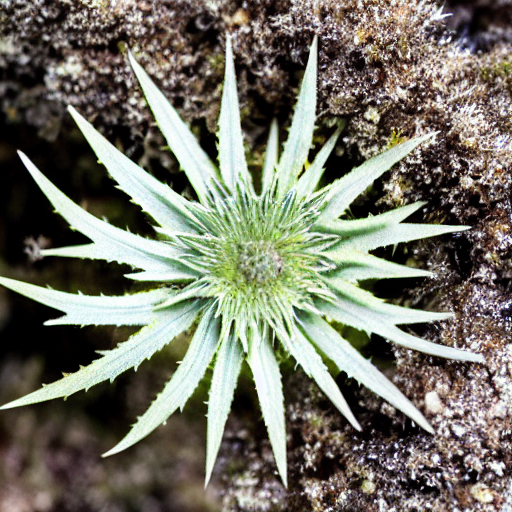} &
\includegraphics[width=0.22\columnwidth]{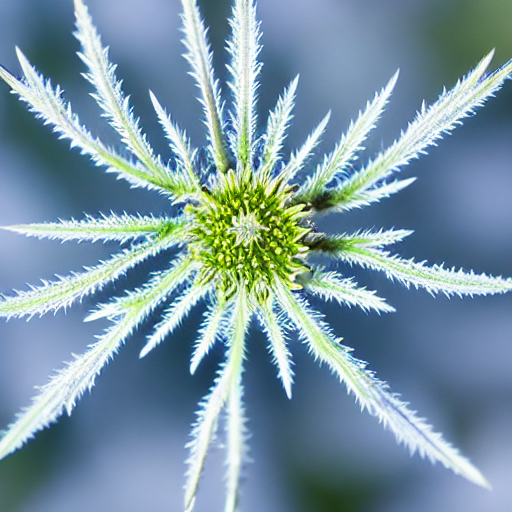} \\
\hline

\multicolumn{4}{|c|}{\textbf{Our Generations for Alpine Sea Holly}} \\

\includegraphics[width=0.22\columnwidth]{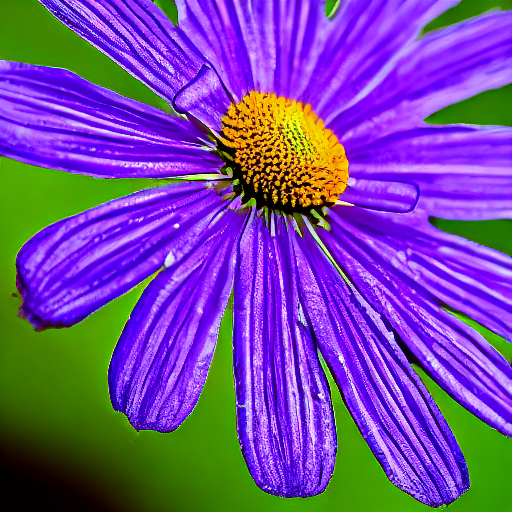} &
\includegraphics[width=0.22\columnwidth]{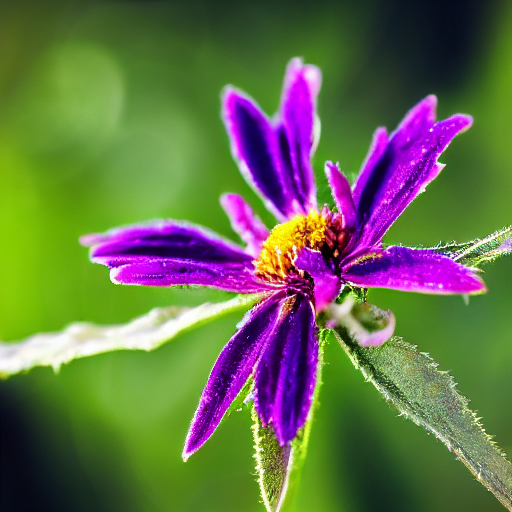} &
\includegraphics[width=0.22\columnwidth]{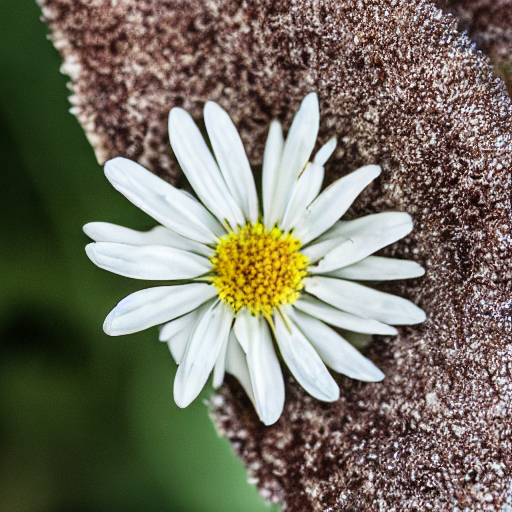} &
\includegraphics[width=0.22\columnwidth]{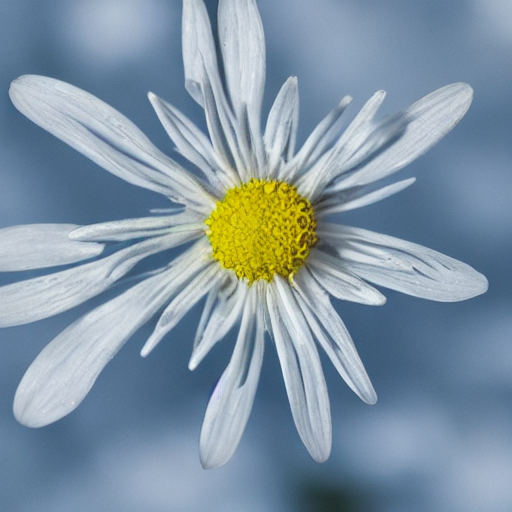} \\
\hline

\multicolumn{4}{|c|}{\textbf{SD1.4 Generations for Predicted Neighbor Class}} \\

\makecell{\textbf{Mexican Aster}} &
\makecell{\textbf{Clematis}} &
\makecell{\textbf{Oxeye Daisy}} &
\makecell{\textbf{Oxeye Daisy}} \\

\includegraphics[width=0.22\columnwidth]{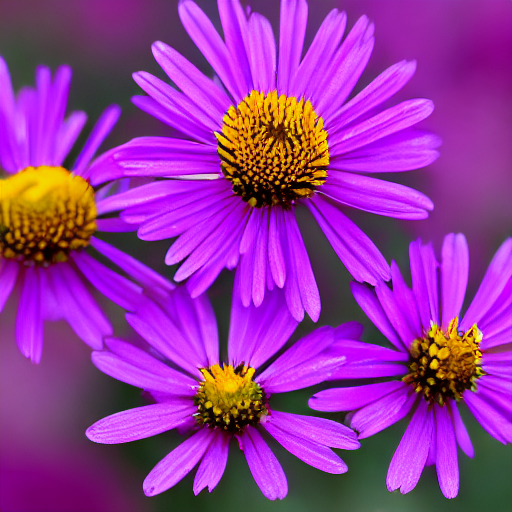} &
\includegraphics[width=0.22\columnwidth]{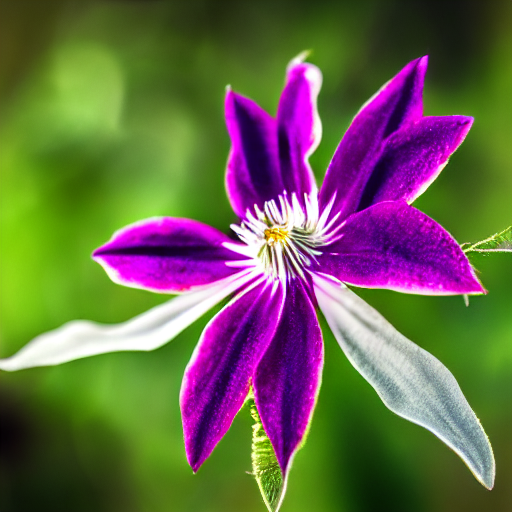} &
\includegraphics[width=0.22\columnwidth]{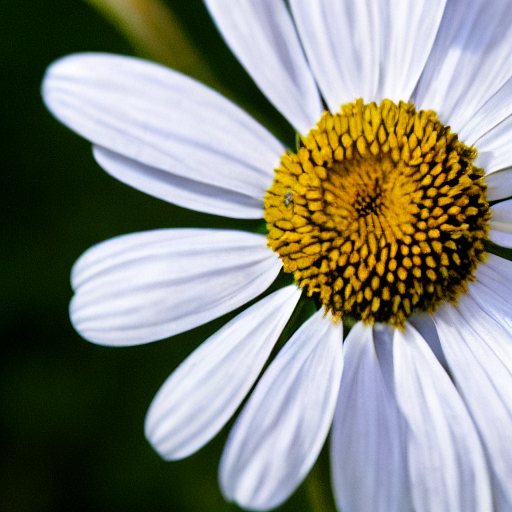} &
\includegraphics[width=0.22\columnwidth]{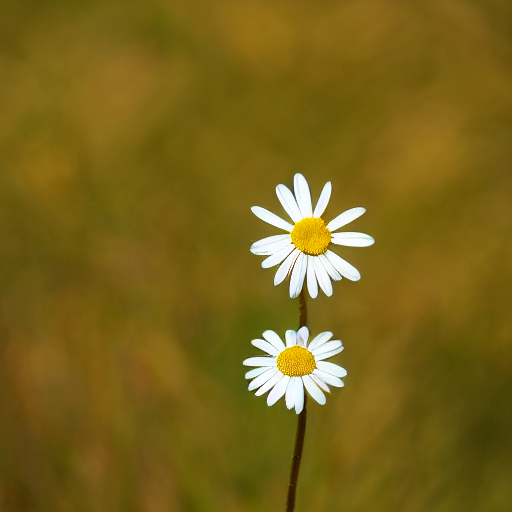} \\
\hline
\end{tabular}%
}

\caption{Concept erasure in a fine-grained setting. The top row shows original SD1.4 generations for \emph{Alpine Sea Holly}, the middle row shows generations after erasure, and the bottom row shows SD1.4 generations conditioned on the classifier-predicted neighboring class.}
\label{fig:consistency}
\end{figure}

Figure~\ref{fig:consistency} illustrates this behavior for the concept \emph{Alpine Sea Holly}. The top row shows original generations from SD1.4, while the middle row shows images generated after erasing the target concept. Although the outputs remain visually similar to the original category, the classifier predicts nearby flower classes such as \emph{Mexican Aster}, \emph{Clematis}, and \emph{Oxeye Daisy}. To further demonstrate this shift, the bottom row shows reference SD1.4 generations conditioned on these predicted labels. The erased outputs visually align more closely with these neighboring classes than with the removed concept. This suggests that the model redirects generation toward semantically adjacent categories.

\section{Adversarial Prompt Robustness}
\label{appendix:adversarial}

We evaluate robustness against adversarially engineered prompts using the
\textit{Ring-A-Bell} red-teaming framework~\cite{tsai2023ring}.  
Ring-A-Bell is a model-agnostic method designed to automatically discover prompts that can bypass safety mechanisms in text-to-image diffusion models by generating prompts that implicitly encode the target concept.

\begin{table}[h]
\centering
\small
\setlength{\tabcolsep}{4pt}
\begin{tabular}{lcccccc}
\toprule
Method & SD-1.4 & SLD & ESD-u & UCE & RECE & Ours \\
\midrule
Ring-A-Bell $\downarrow$
& 83.10 & 66.20 & 69.72 & 33.10 & \textbf{13.38} & \underline{29.6} \\
\bottomrule
\end{tabular}
\caption{Attack success rate under the Ring-A-Bell attack (lower is better).}
\label{tab:ring_a_bell}
\end{table}

Table~\ref{tab:ring_a_bell} reports the attack success rate of different concept removal methods under the Ring-A-Bell attack. Lower values indicate stronger robustness. Our method achieves a low attack success rate and ranks among the baselines, demonstrating improved robustness to adversarial prompt attacks. Although RECE attains slightly higher robustness via closed-form weight updates, it degrades general concept performance.

\section{Neighbor Lists}
\label{appendix:neighbor}

Table~\ref{tab:neighbor} reports the neighbor concepts retrieved by our method for each target concept across the Oxford Flowers, Stanford Dogs, and Celebrity datasets.  
Below we provide additional details on the neighbor mining and ranking procedure.
Given a target concept $c$, we retrieve semantically related concepts from a large external pool $\mathcal{C}_{\text{all}}$ (e.g., Wikipedia titles).  
First, we compute cosine similarity between the embedding of the target concept $x_f \in X_{F_c}$ and candidate embeddings $x_i \in \mathcal{C}_{\text{all}}$:

\[
\cos(x_f, x_i) = \frac{x_f^\top x_i}{\|x_f\|\|x_i\|}.
\]

The top-$k$ most similar concepts form an initial candidate set $\mathcal{C}_k$.

Next, we filter candidates using a RoBERTa-based SVR concreteness predictor~\cite{wartena2024estimating}, keeping only concepts with concreteness score $s_i \ge \tau$.  
To remove rare or overly abstract concepts, we additionally require a minimum popularity threshold $\text{Pop}(c_i) \ge P_{\text{thresh}}$, measured via Wikipedia page views.

\paragraph{Visual CLIP Re-ranking.}
The remaining candidates are re-ranked according to their visual CLIP similarity to the target concept.  
For each concept, we generate $m$ images using a fixed prompt and compute normalized CLIP image embeddings. These embeddings are averaged to obtain a concept prototype $\bar{v}_c$.  
Candidate neighbor concepts are ranked by cosine similarity between their prototypes:

\[
\text{Sim}_{\text{CLIP}}(c,c_i) =
\frac{\bar{v}_c^\top \bar{v}_{c_i}}
{\|\bar{v}_c\|\|\bar{v}_{c_i}\|}.
\]

Unless otherwise specified, we use $m=10$ images per concept.

\begin{figure}[h]
\centering
\includegraphics[width=0.75\columnwidth]{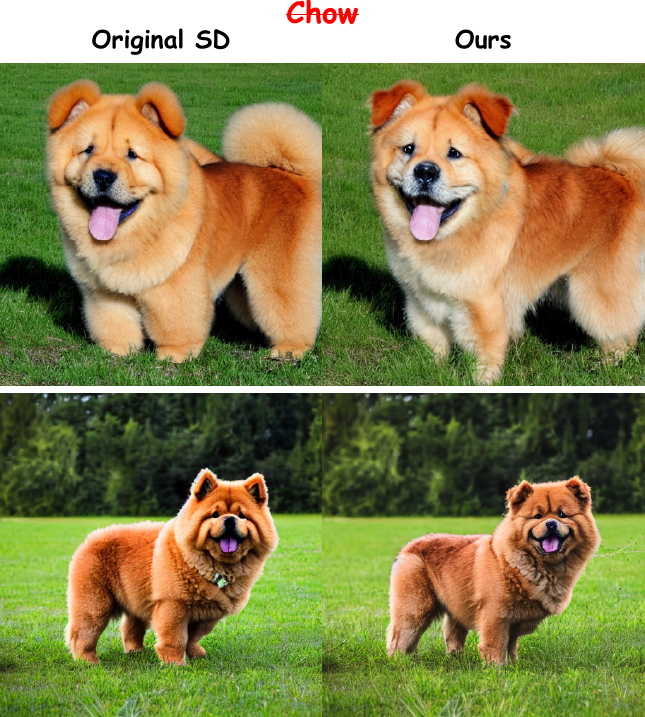}
\caption{
    Qualitative Example of Ineffective Erasure. Due to textual ambiguity, {\methodname} has difficulty identifying an appropriate neighborhood set, resulting in ineffective erasure of the target concept}
    \label{fig:dog_failure}
\end{figure}

\section{Limitations and Failure Cases}

While {\methodname} generally achieves effective erasure without affecting neighboring concepts, it does have limitations. The method depends on both the text-embedding model and the text-to-image model to identify semantically related neighbors. As a result, it can fail when the target concept is ambiguous or poorly represented in either model. This is visible in the multi-concept erasure setting, where a single failure inflated the mean Acc$_t$. For one dog breed, `\textit{Chow}', {\methodname} is unable to identify meaningful neighbors (Figure~\ref{fig:dog_failure}). Instead, it returns the following unrelated concepts:
\texttt{Chi-Chi's, Chi La Sow, Chuño, Zuchu, Wee Kim Wee, Wendy Choo, Shou Zi Chew, Chumlee, Kachhwaha, Lily Chou-Chou}.

These mismatched neighbors cause the projected subspace to be misaligned, which leads to ineffective erasure.

This issue can be reduced by strengthening the underlying models or by disambiguating the target concept. For example, when selecting neighbors for `\textit{Chow (Dog Breed)}', the retrieved neighbors become:
\texttt{Pomeranian dog, Welsh Corgi, Rough Collie, Kangal Shepherd Dog, Caucasian Shepherd Dog, Shikoku dog, Kombai dog, Bernese Mountain Dog, Bichon, Central Asian Shepherd Dog}.

These neighbors are much more reliable erasure.

\section{More Qualitative Example}
\label{appendi:qualitative}
\begin{figure*}[!t]
\centering
\includegraphics[width=\textwidth]{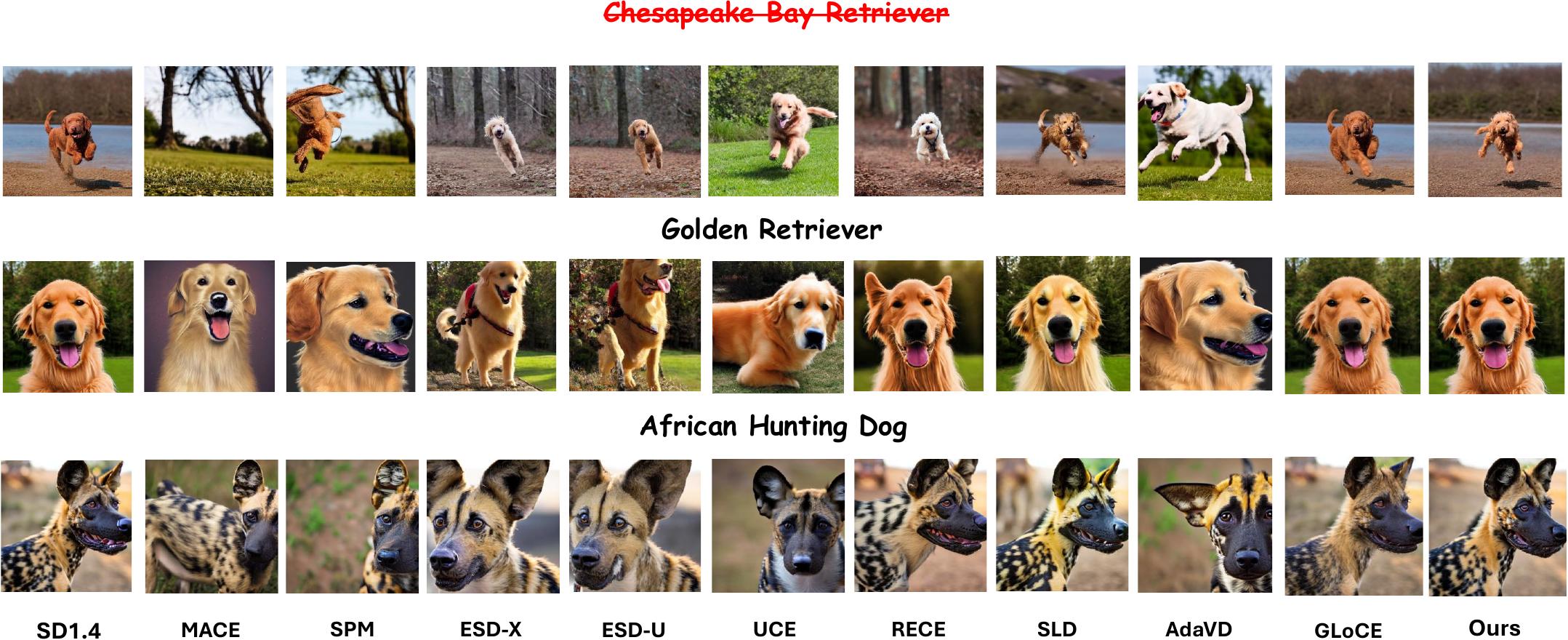}
\caption{
    Further Qualitative Comparisons on the Stanford Dogs dataset. Our {\methodname} can effectively remove the target '\textit{Chesapeake Bay Retriever}' while preserving other dog breeds like '\textit{Golden Retriever}' and '\textit{African Hunting Dog}'.}
    \label{fig:dog_more}
\end{figure*}

\begin{figure*}[!t]
\centering
\includegraphics[width=\textwidth]{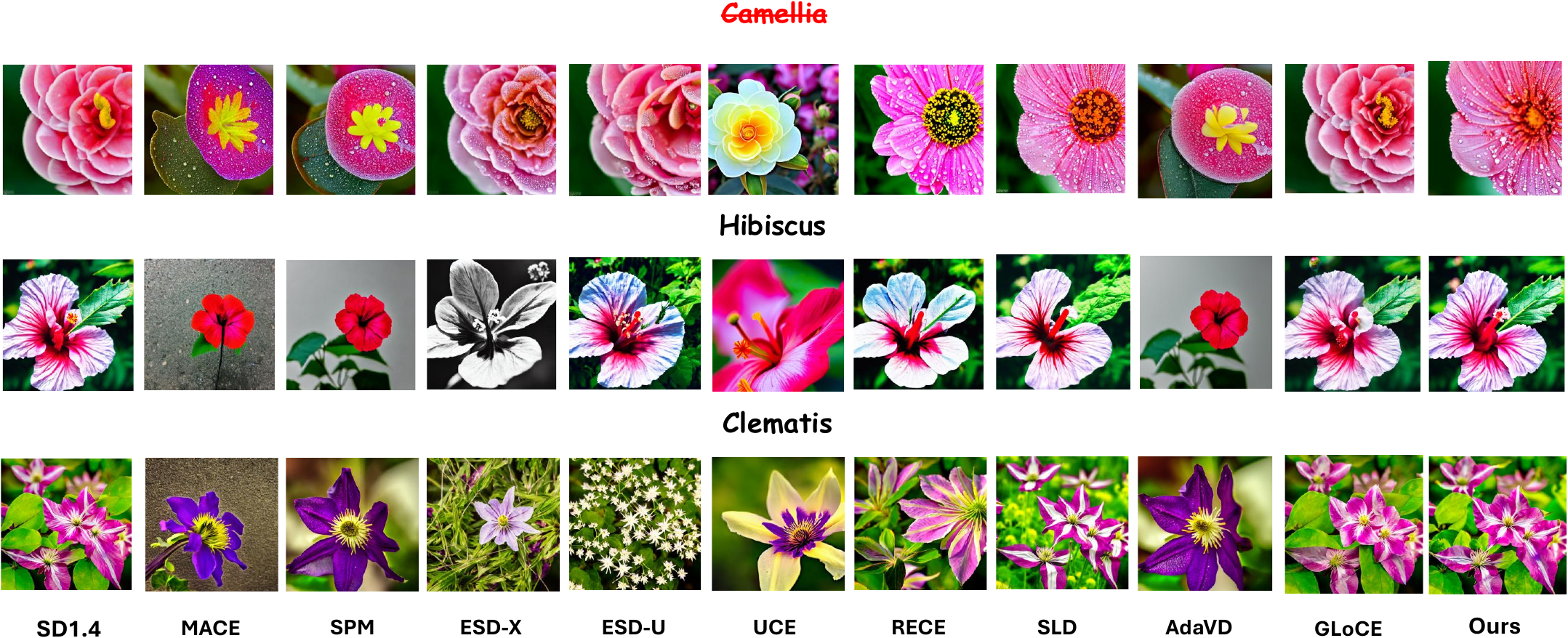}
\caption{
    Further Qualitative Comparisons on the Oxford Flowers dataset. Our {\methodname} can effectively remove the target '\textit{Camellia}' while preserving other flower types like '\textit{Hibiscus}' and '\textit{Clematis}'.}
    \label{fig:flower_more}
\end{figure*}

\begin{figure*}[!t]
\centering
\includegraphics[width=\textwidth]{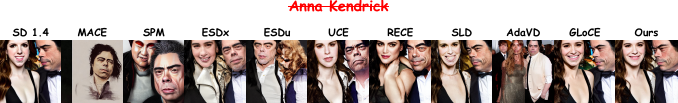}
\caption{
    Further Qualitative Comparisons on the Celebrity dataset. Our {\methodname} can effectively remove the target '\textit{Anna Kendrick}' while preserving other celebrity '\textit{Benicio Del Toro}'.}
    \label{fig:celeb_more}
\end{figure*}

\begin{figure*}[h]
\centering
\includegraphics[width=\textwidth]{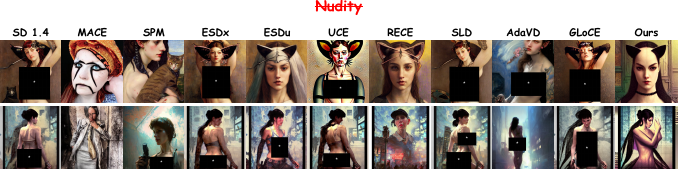}
\caption{
    Further Qualitative examples on I2P dataset for explicit content erasure.}
    \label{fig:nudity_more}
\end{figure*}

\begin{figure*}[h]
\centering
\includegraphics[width=\textwidth]{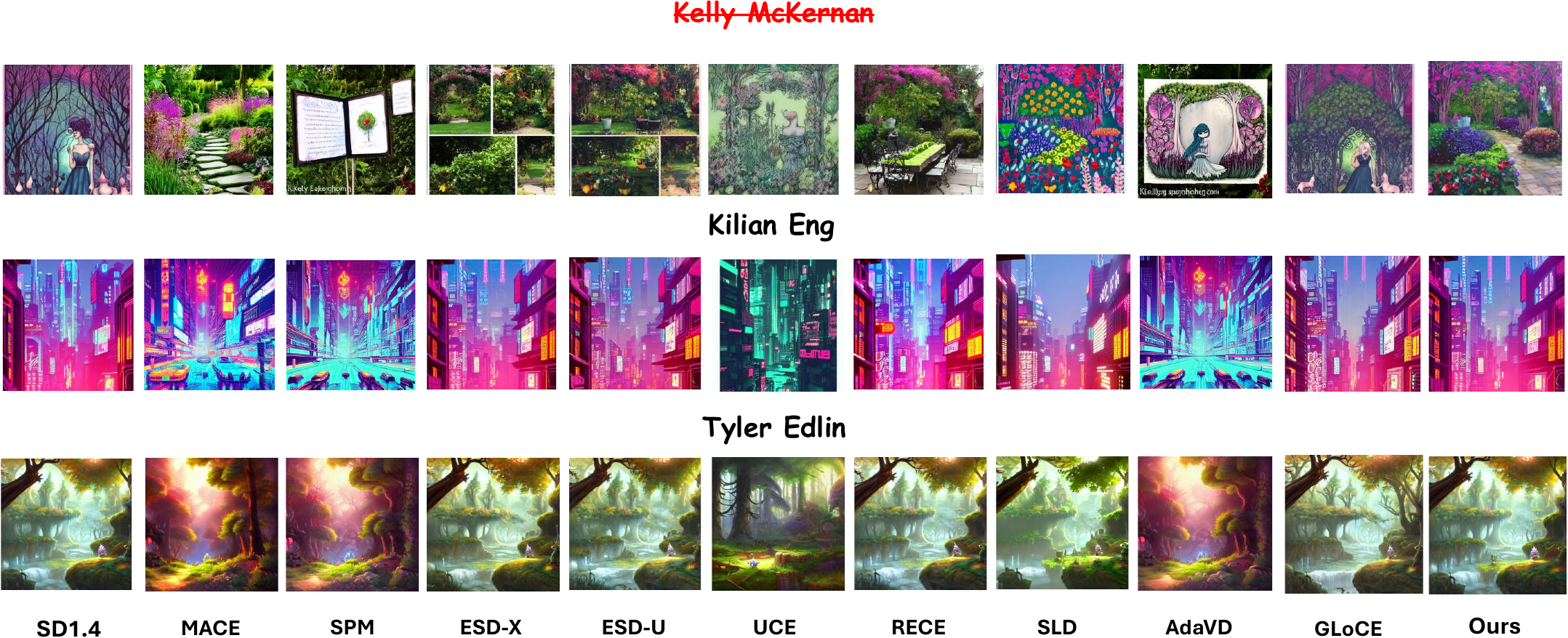}
\caption{
    Further Qualitative Comparisons on the Artistic Dataset. Our {\methodname} can effectively remove the target style '\textit{Kelly Mckernan}' while preserving style like '\textit{Kilian Eng}' and '\textit{Tyler Edlin}'.}
    \label{fig:artistic_more}
\end{figure*}

Across a wide range of datasets, our additional qualitative examples further demonstrate the generality and precision of {\methodname}. As illustrated in Figure \ref{fig:dog_more}, our approach can effectively remove the designated target category in the Stanford Dogs dataset while preserving non-target content. Similarly, Figure \ref{fig:flower_more} shows successful removal of the specified target category in the Oxford Flowers dataset without affecting surrounding classes. In Figure \ref{fig:celeb_more}, our method reliably eliminates the target identity in the celebrity dataset while preserving other individuals. Finally, Figure \ref{fig:nudity_more} provides further results on the I2P dataset, highlighting {\methodname}’s consistent ability to remove explicit content.
Figure \ref{fig:artistic_more} presents qualitative results on artistic data, where {\methodname} removes the target style while maintaining the integrity of other stylistic features.

\end{document}